\documentclass{ecai} 

\usepackage{lipsum}

\usepackage{latexsym}
\usepackage{amssymb}
\usepackage{amsmath}
\usepackage{amsthm}
\usepackage{bm}
\usepackage{booktabs}
\usepackage{enumitem}
\usepackage{graphicx}
\usepackage{subfig}
\usepackage{cuted}
\usepackage{color}
\usepackage{marvosym}

\newtheorem{theorem}{Theorem}
\newtheorem{lemma}{Lemma}
\newtheorem{definition}{Definition}

\newcommand{\BibTeX}{B\kern-.05em{\sc i\kern-.025em b}\kern-.08em\TeX}

\begin{document}


\begin{frontmatter}


\paperid{1012} 


\title{The Role of Depth, Width, and Tree Size in Expressiveness of Deep Forest}


\author[1,2]
{Shen-Huan Lyu\footnote[\dagger]{Equal contribution.}}

\author[3,4]
{Jin-Hui Wu\footnote[\dagger]{}}

\author[3,4]
{Qin-Cheng Zheng}

\author[3]
{Baoliu Ye\footnote[\ddagger]{Corresponding author.}}

\address[1]{Key Laboratory of Water Big Data Technology of Ministry of Water Resources, Hohai University, China}

\address[2]{College of Computer Science and Software Engineering, Hohai University, China}

\address[3]{National Key Laboratory for Novel Software Technology, Nanjing University, China}

\address[4]{School of Artificial Intelligence, Nanjing University, China}

\address{\texttt{lvsh@hhu.edu.cn} \quad \texttt{\{wujh,zhengqc\}@lamda.nju.edu.cn} \quad \texttt{yebl@nju.edu.cn}}



\begin{abstract}
Random forests are classical ensemble algorithms that construct multiple randomized decision trees and aggregate their predictions using naive averaging.
\citet{zhou2019deep} further propose a deep forest algorithm with multi-layer forests, which outperforms random forests in various tasks.
The performance of deep forests is related to three hyperparameters in practice: depth, width, and tree size, but little has been known about its theoretical explanation. 
This work provides the first upper and lower bounds on the approximation complexity of deep forests concerning the three hyperparameters.
Our results confirm the distinctive role of depth, which can exponentially enhance the expressiveness of deep forests compared with width and tree size.
Experiments confirm the theoretical findings.
\end{abstract}

\end{frontmatter}

\section{Introduction}

Random forests~\cite{breiman2001random} and neural networks~\cite{anthony1999neural} are regarded as two contrasting approaches to learning. The former is considered more suitable for modeling categorical and mixed data, such as medical diagnosis analysis \citep{basu2018iterative} and financial anomaly detection \citep{afriyie2023supervised}. In contrast, the latter is better suited for modeling numerical data, such as computer vision \citep{krizhevsky2017imagenet} and natural language processing \citep{devlin2019bert}. Random forests are favored for the tree-based intuitive inference, which makes them easier for users to understand and utilize \citep{breiman2001random}. On the other hand, neural networks are renowned for their distinguishing performance when dealing with complex data, even though they are often perceived as opaque black-box models that are difficult to comprehend \citep{murdoch2019definitions}.


Recently, deep learning \citep{lecun2015deep} has significantly improved the expressiveness and performance of neural networks. Expressiveness of a model implies its hypothesis class complexity, and higher expressiveness implies higher approximation efficiency. Numerous studies have demonstrated that, within the neural network architecture, depth plays a crucial role in efficiently representing complex data, surpassing width exponentially \citep{daniely2017depth,eldan2016power,hu2022analysis,vardi2022width}. Neural networks rely on gradient propagation for training, but their performance on many categorical datasets is often inferior to that of traditional tree-based learning algorithms.
Therefore, many real-world tasks require algorithms composed of non-differentiable modules, such as random forests \citep{breiman2001random,geurts2006extremely}, GBDTs \citep{chen2016xgboost,ke2017light}, etc.

By realizing that the essence of deep learning lies in \textit{layer-by-layer processing}, \textit{in-model feature transformation}, and \textit{sufficient model complexity},
\citet{zhou2019deep} propose the first non-differentiable deep model based on decision trees, known as deep forest. In practice, deep forests outperform various ensemble algorithms based on decision trees and have been involved in real applications such as biomedicine \citep{guo2018bcdforest}, smart water management \citep{liu2019deep}, and financial risk assessment \citep{zhang2019distributed}, etc. Various adaptations of deep forests have excelled in diverse learning scenarios \citep{utkin2019discriminative,wang2020learning,yang2020multilabel}, and efforts are ongoing to improve the learning efficiency and reduce the computational cost for large-scale deep forests~\citep{lyu2022region,pang2022improving,zhou2019deephash}.

The success of deep forests in practice has attracted attention for its theoretical analysis. Regarding \textit{layer-by-layer processing}, \citet{lyu2019refined} prove that deep forests can optimize sample margin distributions layer by layer, thereby alleviating the overfitting risk. 
The axis-aligned structure of decision trees is considered to prove that \textit{layer-by-layer processing} significantly improves the consistency rate of random forests \citep{arnould2021analyzing,lyu2022depth}. 
In terms of \textit{in-model feature transformation}, a line of work shows that new features based on predictions can easily cause overfitting risk, and propose a novel feature representation method based on decision rules \citep{chen2021improving,lyu2022region,pang2022improving}. However, there is still a lack of theoretical explanation for \textit{sufficient model complexity}.

\begin{figure}[t]
    \vspace{-0.1cm}
    \centering
    \includegraphics[width=\columnwidth]{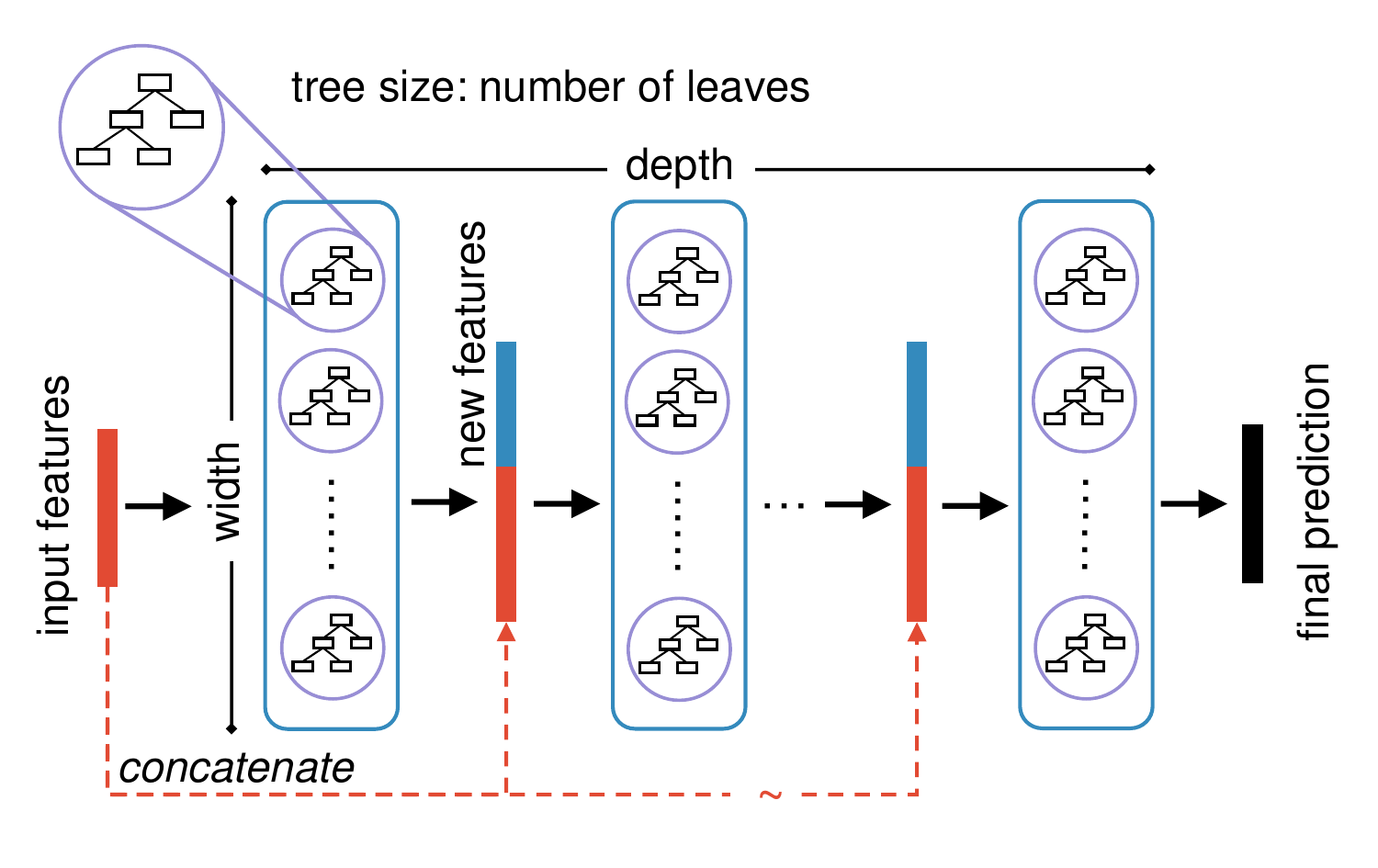}
    \caption{Illustration of the deep forest architecture.}
    \label{fig:deepforest}
    \vspace{0.2cm}
\end{figure}

As shown in Figure~\ref{fig:deepforest}, a single decision tree serves as the unit of a deep forest, then the number of parameters in a decision tree is referred to as the `tree size’. Each layer of deep forests consists of a certain number of decision trees. This is known as the `width’ of deep forests. Additionally, predictions from each layer are transmitted to the next layer as augmented features. This iterative process continues for a specified number of layers, which is referred to as the `depth’ of deep forests. These three hyperparameters impact the complexity of deep forests in practice. Notably, the depth distinguishes the deep forests from individual decision trees and random forests. However, the theoretical advantage of depth remains unclear.

\paragraph{Contributions.} In this paper, we show the advantages of depth over width and tree size from the perspective of expressiveness (refer to approximation complexity in Definition~\ref{def:approximationcomplexity}), focusing on the generalized parity functions and a simplified architecture of deep forest.
The main contributions can be summarized as follows:
\begin{itemize}[itemsep=0pt,topsep=0pt]
    \item We show that depth is more powerful than tree size and width from the aspect of expressiveness. Specifically, we prove the advantage of depth over tree size by separation results of expressing parity functions and the worst-case guarantees, as shown in Theorems~\ref{thm: depth vs tree size, separation} and~\ref{thm: depth vs tree size, worst case guarantee}, respectively. We show that depth can be more efficient than width by separation results, as shown in Theorem~\ref{thm: depth vs width, separation without error}.
    \item We further demonstrate the power of depth from the aspect of learning in experiments. Specifically, we construct a product distribution to learn parity functions and verify our theoretical findings via simulation. Real-world experiments also support the efficiency of depth in deep forests.
\end{itemize}

\paragraph{Organization.}

The rest of this work is organized as follows: Section~\ref{sec:relatedwork} introduces related work.
Section~\ref{sec:preliminaries} provides basic notations and definitions. 
Section~\ref{sec:mainresults} compares depth, width, and tree size in deep forests via expressiveness.
Section~\ref{sec:experiments} presents experiments to confirm the theoretical results.
Section~\ref{sec:conclusion} concludes with future work. 

\section{Related work}\label{sec:relatedwork}

Random forests aggregate multiple decision trees along the width dimension to improve performance. Therefore, theoretical properties of width have attracted significant interest~\citep{biau2012analysis,denil2014narrowing}. \citet{breiman2001random} offers an upper bound on the generalization error of random forests in terms of correlation and accuracy of individual trees. In recent years, various theoretical works \citep{biau2008consistency,biau2012analysis,ishwaran2010consistency,genuer2012variance,zhu2015reinforcement} have been performed, analyzing the consistency of various simplified forests, and moving ever closer to practice.
\citet{scornet2015consistency} prove the first $\mathbb{L}_2^2$-consistency of Breiman's original random forests with CART-split criterion \citep{breiman1984classification}.
\citet{wang2023stability} show that a larger width can enhance the stability of random forests. \citet{curth2024why} present the adaptive smoothing behavior of random forests over decision trees.
Additionally, tree size plays a crucial role in random forests. \citet{scornet2015consistency} prove that limiting the size of decision trees can lead to insufficient diversity, which slows down the consistency rate.

Although deep forests achieve satisfactory performance in many tasks, there lacks theoretical discussions on how the depth, width, and tree size impact the capability of deep forests. Approximation complexity reflects the capability of approximating certain function classes and is widely studied in neural networks to compare the impact of different hyperparameters on approximation efficiency. Many attractive conclusions are derived from analyzing approximation complexity of neural networks, such as depth is more powerful than width and activation complexity~\cite{montufar2014number,eldan2016power,telgarsky2016benefits,daniely2017depth,vardi2022width,goujon2024number}, and complex-valued networks are more powerful than real-valued networks~\cite{zhang2022towards,wu2023theoretical}. This paper takes the first step towards investigating approximation capability for tree-based models and proves that depth can be more powerful than width and tree size in achieving sufficient model complexity using fewer number of leaves.

\section{Preliminaries}\label{sec:preliminaries}

Let $\mathcal{X} = \mathcal{X}_1 \times \mathcal{X}_2 \times \dots \times \mathcal{X}_n$ denote the $n$-dimensional discrete input space, where $\mathcal{X}_1 , \mathcal{X}_2 , \dots , \mathcal{X}_n \subset \mathbb{R}$ are finite sets. For simplicity, these finite sets are supposed to share the same domain, i.e., $\mathcal{X}_1 = \dots = \mathcal{X}_n = [p] := \{ 1 , 2 , \dots , p \}$. We consider the binary classification problem in this paper, i.e., the target concept $c$ is a mapping from the input space $\mathcal{X}$ to the output space $\mathcal{Y} = \{ -1 , +1 \}$. Let $\bm{v}_i$ indicate the $i$-the coordinate of the vector $\bm{v}$, and $\dim(\cdot)$ denotes the dimension of a vector or a space.

This work focus on the expressiveness of decision trees, forests, and deep trees, where trees in forests and deep trees have a restricted number of leaves. These three models have one unrestricted and two restricted hyperparameters, as shown in Table~\ref{tab: hyperparameters of three models}. We proceed to provide formal definitions for the size of the three mentioned models. Let $\mathcal{P}_\mathrm{T}$ denote the parameter space of binary decision trees, which satisfies the following requirements: i) $\mathcal{Y} \subset \mathcal{P}_\mathrm{T}$; ii) $(i ; x_i ; \Theta_L ; \Theta_R) \in \mathcal{P}_\mathrm{T}$ holds for any $i \in [\dim(\mathcal{X})]$, $x_i \in \mathcal{X}_i$, and $\Theta_L , \Theta_R \in \mathcal{P}_\mathrm{T}$. Then any parameter $\Theta \in \mathcal{P}_\mathrm{T}$ determines a decision tree as follows
\begin{equation*}
  h_\Theta (\bm{x}) = \left\{
    \begin{array}{ll}
      \Theta_1 , & \dim(\Theta) = 1 , \\
      h_{\Theta_L} (\bm{x}) , & \dim(\Theta) > 1 \text{~and~} \bm{x}_{\Theta_1} \leqslant \Theta_2 , \\
      h_{\Theta_R} (\bm{x}) , & \dim(\Theta) > 1 \text{~and~} \bm{x}_{\Theta_1} > \Theta_2 .
    \end{array}
  \right.
\end{equation*}
All such trees form the hypothesis space of decision trees
\begin{equation*}
  \mathcal{H}_\mathrm{T} (\mathcal{X} , \mathcal{Y})
  = \{ h_\Theta : \mathcal{X} \rightarrow \mathcal{Y} \mid \Theta \in \mathcal{P}_\mathrm{T} \} .
\end{equation*}
\begin{table}[h]
\vspace{-0.4cm}
    \centering
    \caption{Hyperparameters of tree-based models.}
    \vspace{0.2cm}
    \label{tab: hyperparameters of three models}
    \begin{tabular}{cccc}
        \toprule
        & Tree Size & Width & Depth \\
        \midrule
        Tree & Unrestricted & Restricted & Restricted \\
        Forest & Restricted & Unrestricted & Restricted \\
        Deep Tree & Restricted & Restricted & Unrestricted \\
        \bottomrule
    \end{tabular}
\end{table}

Define $\dim(h_\Theta) = \dim(\Theta)$ as the size of a tree $h_\Theta \in \mathcal{H}_\mathrm{T}$. A decision tree with $m$ parent nodes satisfies $\dim (h_{\Theta}) = 3m+1$. We are also interested in decision trees of restricted size. More specially, the maximal number of parent nodes is set as twice the input size for simplicity, or equivalently,
\begin{equation*}
  \mathcal{H}_{\mathrm{T}^{\circ}} (\mathcal{X} , \mathcal{Y})
  = \{ h_\Theta \in \mathcal{H}_\mathrm{T} \mid \dim (h_\Theta) \leqslant 6 \dim (\mathcal{X}) + 1 \} .
\end{equation*}
Let $M(\bm{v})$ represent the mode (or majority vote) of the vector $\bm{v}$. When the mode is not unique, we set $M(\bm{v})$ by uniformly randomly choosing one from all modes. We focus on the forests composed of trees with restricted size, whose hypothesis space is
\begin{equation*}
  \begin{aligned}
    \mathcal{H}_\mathrm{F} (\mathcal{X} , \mathcal{Y})
    = \{ h \mid &~ \exists~ N \in \mathbb{N}^+ , h_{\Theta_1} , \dots , h_{\Theta_N} \in \mathcal{H}_{\mathrm{T}^{\circ}}( \mathcal{X} , \mathcal{Y} ) , \\
    &~ \text{s.t.~} h(\bm{x}) = M( h_{\Theta_1}(\bm{x}) , \dots , h_{\Theta_N}(\bm{x})) \} ,
  \end{aligned}
\end{equation*}
and $\dim(h) = \dim(h_{\Theta_1}) + \dots + \dim(h_{\Theta_N})$ is the size of a forest $h \in \mathcal{H}_\mathrm{F}$. Let $f : \mathcal{X} \rightarrow \mathcal{Y}$ and $g : \mathcal{X} \times \mathcal{Y} \rightarrow \mathcal{Z}$ be two mappings. Define the cascade composition of $g$ and $f$ as $g \oplus f= g(x,f(x))$. Then the hypothesis space of restricted-tree-size deep trees is
\begin{equation*}
  \begin{aligned}
    \mathcal{H}_\mathrm{DT} (\mathcal{X} , \mathcal{Y})
    = \{ h \mid &~ \exists~ D \in \mathbb{N}^+ , h_{\Theta_1} \in \mathcal{H}_{\mathrm{T}^{\circ}} (\mathcal{X} , \mathcal{Y}) , \\
    &~ h_{\Theta_2} , \dots , h_{\Theta_D} \in \mathcal{H}_{\mathrm{T}^{\circ}} (\mathcal{X} \times \mathcal{Y} , \mathcal{Y}) , \\
    &~ \text{s.t.~} h = h_{\Theta_D} \oplus \dots \oplus h_{\Theta_1} \} ,
  \end{aligned}
\end{equation*}
and $\dim(h) = \dim(h_{\Theta_1}) + \dots + \dim(h_{\Theta_D})$ is the size of a deep tree $h \in \mathcal{H}_\mathrm{DT}$. We then introduce approximation complexity, which is used to compare the expressiveness of different models.

\begin{definition}\label{def:approximationcomplexity}
  Let $\mathcal{H}$ denote a hypothesis space, $c$ represents a target concept, $\epsilon \in [0,1]$ indicates the approximation error, and $\mathcal{D}$ is a distribution on the input space $\mathcal{X}$. Define the approximation complexity $\mathcal{C} ( \mathcal{H} , c , \mathcal{D} , \epsilon )$ as
  \begin{equation*}
    \mathcal{C} ( \mathcal{H} , c , \mathcal{D} , \epsilon )
    = \min_{h \in \mathcal{H}} \{ \dim(h) \mid \Pr_{\bm{x} \sim \mathcal{D}} [ h(\bm{x}) \neq c(\bm{x}) ] \leqslant \epsilon \} .
  \end{equation*}
\end{definition}

The approximation complexity $\mathcal{C} ( \mathcal{H} , c , \mathcal{D} , \epsilon )$ measures the number of parameters required to express target concept $c$ using models from the hypothesis space $\mathcal{H}$, under the distribution $\mathcal{D}$ and within an error tolerance $\epsilon$. It shows the expression capability of the hypothesis space, which is the premise of promising performance. A summary of mathematical symbols is provided in Appendix~\ref{app: mathematical symbols}.

\section{Main results}\label{sec:mainresults}

In this section, we demonstrate theoretical advantages of depth over tree size and width from the perspective of approximation complexity. In Section~\ref{subsec: generalized parity functions}, we provide the definition and several useful properties of generalized parity functions. Sections~\ref{sec: depth vs tree size} and~\ref{sec: depth vs width} show the advantage of depth over tree size and width, respectively, when approximating generalized parity functions.

\subsection{Generalized parity functions}
\label{subsec: generalized parity functions}

We first define the generalized parity functions
\begin{equation*}
    c (\bm{x})
    = (-1)^{\Vert \bm{x} \Vert_1}
    \quad \text{for} \quad
    \bm{x} \in [p]^n ,
\end{equation*}
which degenerates to the parity function when $p=2$. We proceed to introduce the notion of label-connected sets, which is used to characterize the complexity of parity functions.

\begin{definition}
  \label{def: label connected}
  Let $\bm{x} , \bm{y} \in \mathcal{X}$ denote two input vectors, $r>0$ is a positive real number, and $f : \mathcal{X} \rightarrow \{ -1 , 1 \}$ represents a classification mapping. Vectors $\bm{x}$ and $\bm{y}$ are $(r,f)$-label-connected if there exist $n$ input vectors $\bm{z}_1 = \bm{x} , \bm{z}_2 , \dots , \bm{z}_n = \bm{y} \in \mathcal{X}$, such that $\Vert \bm{z}_{i+1} - \bm{z}_i \Vert_1 \leqslant r$ and $f(\bm{z}_{i+1}) = f(\bm{z}_i)$ holds for any $i \in [n-1]$. The $(r,f)$-label-connected set of vector $\bm{x}$ is defined as $C_{r,f}(\bm{x}) = \{ \bm{y} \in \mathcal{X} \mid \bm{x} \text{~and~} \bm{y} \text{~are~} (r,f)\text{-label connected} \}$.
\end{definition}

We mostly focus on the $(1,f)$-label-connected sets throughout this paper. When the input dimension is $2$, imagine the input space $\mathcal{X}$ as a chess board, the points with the same label to $\bm{x}$ as pools, and the other points as mountains. Two adjacent pools are connected and otherwise severed. Then the $(1,f)$-label-connected set of $\bm{x}$ depicts the water area containing $\bm{x}$. From this intuitive explanation, it is observed that either two points share the same water area (label-connected set) or their water areas are disjoint. We formally claim this property for general label-connected sets in the following lemma and provide rigorous proof in Appendix~\ref{app: complete proof for parity functions and product distributions}.

\begin{lemma}
  \label{lem: coincident or disjoint label connected sets}
  Let $\bm{x} , \bm{y} \in \mathcal{X}$ denote two input vectors, $r>0$ is a positive real number, and $f \in \mathcal{F}_\mathcal{X}$ represents a classification mapping. Then either $C_{r,f}(\bm{x}) = C_{r,f}(\bm{y})$ or $C_{r,f}(\bm{x}) \cap C_{r,f}(\bm{y}) = \varnothing$.
\end{lemma}

Lemma~\ref{lem: coincident or disjoint label connected sets} shows that the label-connected sets of two inputs are either the same or disjoint. Then label-connected sets of all inputs form a partition of the input space by the connectivity of labels. The cardinality of the partition reflects the complexity of the function since each element of the partition corresponds to a constant piece of the function. For the simplest constant function, the cardinality equals 1. For parity functions, the function values are different on any two adjacent inputs, leading to an exponential cardinality when considering $(1,f)$-label-connected sets. The following lemma formally presents the cardinality for parity functions and is proved in Appendix~\ref{app: complete proof for parity functions and product distributions}.

\begin{lemma}
  \label{lem: properties of parity functions}
  Let $\mathcal{X} = [p]^n$ be the input space, and $c(\bm{x}) = (-1)^{\Vert \bm{x} \Vert_1}$ is the generalized parity function with $\bm{x} \in \mathcal{X}$. Then we have
  \begin{enumerate}[itemsep=1pt,topsep=1pt]
      \item $| \{ C_{1,c} (\bm{x}) \} | = p^n$.
      \item $\big| | \{ C_{1,c} (\bm{x}) \mid c(\bm{x}) = 1 \} | - | \{ C_{1,c} (\bm{x}) \mid c(\bm{x}) = -1 \} | \big| \leqslant 1$.
  \end{enumerate}
\end{lemma}

Lemma~\ref{lem: properties of parity functions} demonstrates two properties of parity functions. The first one implies that the cardinality of all label-connected sets is exponential of input dimension $n$, and the second one shows that the cardinality of positive label-connected sets is almost the same as that of negative ones. Thus, a parity function is a piecewise constant function with exponential pieces, and both positive and negative pieces are exponential. This makes parity functions hard to approximate using piecewise constant functions with polynomial pieces, including decision trees with polynomial leaves. We formally present the hardness of approximation in the following lemma and provide detailed proof in Appendix~\ref{app: complete proof for parity functions and product distributions}.

\begin{lemma}
  \label{lem: parity function and decision tree}
  Let $h_\mathrm{T} \in \mathcal{H}_\mathrm{T}$ represent a decision tree with $L$ leaves, $c$ denotes the parity function. Define $\mathcal{E} (h_\mathrm{T} , c) = \{ \bm{x} \in \mathcal{X} \mid h_\mathrm{T} (\bm{x}) \neq c (\bm{x}) \}$ and $\mathcal{P} (h_\mathrm{T} , c) = \{ \bm{x} \in \mathcal{X} \mid h_\mathrm{T} (\bm{x}) = c (\bm{x}) \}$ as the error set and proper set of the decision tree $h_\mathrm{T}$, respectively. Then we have $| \mathcal{E} (h_\mathrm{T} , c) | \geqslant ( p^n - L ) / 2$ and $| \mathcal{P} (h_\mathrm{T} , c) | \leqslant ( p^n + L ) / 2$.
\end{lemma}

Lemma~\ref{lem: parity function and decision tree} provides a lower bound of the cardinality of misclassified inputs when approximating parity functions using decision trees. As long as the number of leaves in a decision tree is polynomial with respect to the input dimension, the lower bound is dominated by $\Omega (p^n)$. Thus, decision trees suffer an $\Omega (1)$ classification error under the uniform distribution unless the number of leaves is exponential.

\subsection{Depth is more powerful than tree size}
\label{sec: depth vs tree size}

\begin{figure}[t]
  \subfloat[Step 1. Split the plane into strips, where the same pattern have the same pseudo label.]{
    \label{fig: construction of deep tree, step 1, depth vs tree size, separation}
    \includegraphics[width=0.43\linewidth]{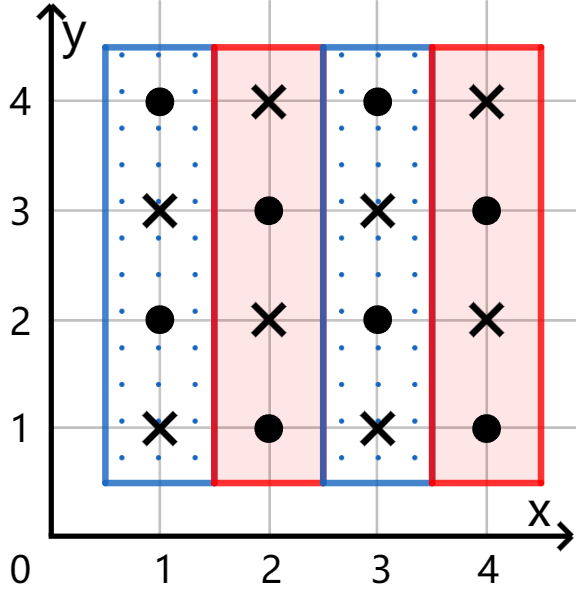}
  }
  \hspace{0.04\linewidth}
  \subfloat[Step 2. Assign final labels after gathering strips with the same pattern using the pseudo labels.]{
    \label{fig: construction of deep tree, step 2, depth vs tree size, separation}
    \includegraphics[width=0.43\linewidth]{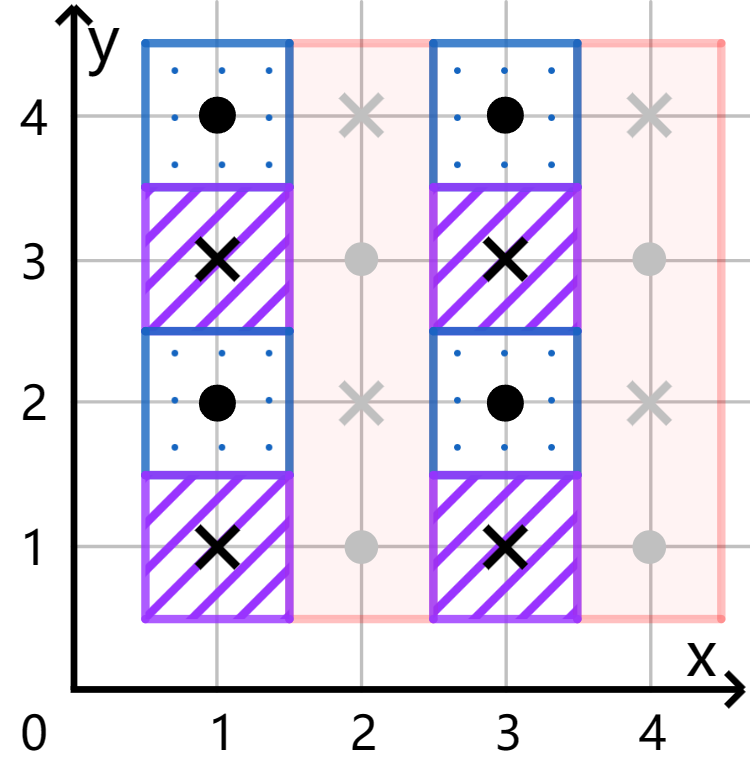}
  }
  \caption{A $2$-dimensional demonstration of the construction of the deep tree expressing the parity function. Circles and crosses at integral points are positive and negative classes, respectively. Rectangles indicate tree leaves.}
  \vspace{0.5cm}
\end{figure}

In this subsection, we compare the efficiency of depth and tree size, beginning with the advantage of depth.

\begin{theorem}
  \label{thm: depth vs tree size, separation}
  For any input space $\mathcal{X}$, there exist a concept $c$ and a distribution $\mathcal{D}$ over $\mathcal{X}$, such that
  \begin{equation*}
    \mathcal{C} ( \mathcal{H}_\mathrm{DT} , c , \mathcal{D} , \epsilon ) 
    \leqslant 10 p n
    \quad \text{and} \quad
    \mathcal{C} ( \mathcal{H}_\mathrm{T} , c , \mathcal{D} , \epsilon ) 
    \geqslant p^n / 2 
  \end{equation*}
  holds for any $\epsilon \in [ 0 , 1/4 ]$.
\end{theorem}

Theorem~\ref{thm: depth vs tree size, separation} shows that there exists a classification mapping such that restricting the depth requires increasing the tree size polynomially (with respect to the feature complexity $p$) or exponentially (with respect to the input dimension $n$). This indicates the efficiency advantage of depth over tree size in approximating particular functions.

The key observation of proof is that each leaf of a decision tree corresponds to a continuous area with the same labels, while each leaf of a deep tree may assign the same label to many disjoint areas. This phenomenon motivates us to construct the parity function, i.e., $c(\bm{x}) = (-1)^{\Vert \bm{x} \Vert_1}$, which maximizes the number of disjoint areas since any two points with the same label are not connected.

For deep trees, the upper bound is proven by construction. Take the $2$-dimensional case as an example. As shown in Figure~\ref{fig: construction of deep tree, step 1, depth vs tree size, separation}, the first part of the deep tree splits the plane into strips using the first coordinate of input. There are only two patterns among all strips since the target concept $c$ is the parity function. The leaves of the first part assign pseudo labels to these strips according to their patterns. Then as shown in Figure~\ref{fig: construction of deep tree, step 2, depth vs tree size, separation}, the second part utilizes the pseudo labels to gather strips with the same pattern and assigns the final labels. Since strips are gathered according to their patterns, each leaf of the second part can label half a line of points, which reduces the complexity dramatically. This intuitive construction can be directly extended to arbitrary input dimension and feature complexity, which leads to a deep tree with $n$ parts and $O(p)$ leaves in each part, i.e., the approximation complexity is $O(pn)$.
\begin{figure}[t]
  \subfloat[Case 1: One more correct point than mistaken points.]{
    \label{fig: leaf of tree, more than 1}
    \includegraphics[width=0.38\linewidth]{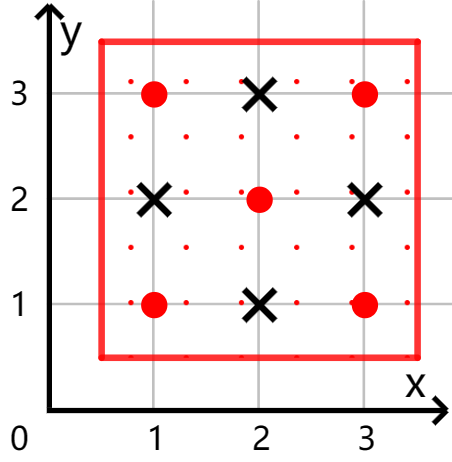}
  }
  \hspace{0.04\linewidth}
  \subfloat[Case 2: The same number of correct points as mistaken points.]{
    \label{fig: leaf of tree, same}
    \includegraphics[width=0.48\linewidth]{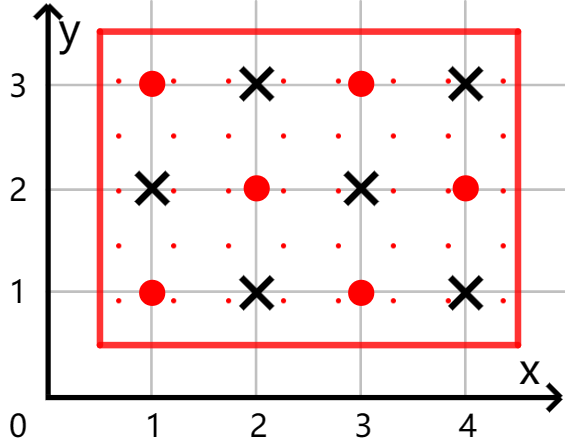}
  }
  \caption{A $2$-dimensional demonstration of the relationship between the number of correct points and the number of mistaken points in a tree leaf, where rectangles, circles, and crosses represent tree leaves, correct points, and mistaken points, respectively.}
  \label{fig: leaf of tree}
  \vspace{0.5cm}
\end{figure}

For decision trees expressing the parity function, a basic observation is that the number of correctly labeled points cannot exceed that of mistakenly labeled in any leaf plus $1$. As shown in Figure~\ref{fig: leaf of tree, more than 1}, the number of correctly labeled points is one more than that of mistakenly labeled points when all widths of the leaf are odd numbers and the leaf receives the dominant label. In the other case as shown in Figure~\ref{fig: leaf of tree, same}, the number of correctly labeled points equals that of mistakenly labeled points when some width of the leaf is an even number. Thus, a decision tree must grow one more leaf to increase the number of correctly labeled points by $1$, which only promotes the accuracy by $1 / p^n$. Therefore, a decision tree requires at least $\Omega (p^n)$ leaves to enhance the accuracy by a constant.

We then study the dual problem of Theorem~\ref{thm: depth vs tree size, separation}, i.e., does decision tree possess an exponential efficiency advantage over deep trees when expressing suitable concepts? The next theorem provides a negative answer for this question.

\begin{theorem}
  \label{thm: depth vs tree size, worst case guarantee}
  For any input space $\mathcal{X}$, any concept $c$, any distribution $\mathcal{D}$ over $\mathcal{X}$, and any $\epsilon \in [0,1]$, one has
  \begin{equation*}
    \mathcal{C} ( \mathcal{H}_\mathrm{DT} , c , \mathcal{D} , \epsilon )
    \leqslant (4n+1) \mathcal{C} ( \mathcal{H}_\mathrm{T} , c , \mathcal{D} , \epsilon ) .
  \end{equation*}
\end{theorem}

Theorem~\ref{thm: depth vs tree size, worst case guarantee} provides the worst-case guarantee for deep trees by showing that for all classification mappings, the approximation complexity of deep trees is no more than that of decision trees multiplying a factor linear in the input dimension $n$. Although decision trees might prevail over deep trees, the advantage of decision trees cannot transcend an upper bound linear in $n$, which is exponentially smaller than the superiority of deep trees over decision trees as demonstrated in Theorem~\ref{thm: depth vs tree size, separation}. This tremendous gap between exponentiality and linearity indicates that deep trees outperform decision trees consistently from the perspective of approximation complexity.

\begin{figure}[t]
\centering
  \subfloat[Output of a decision tree.]{
    \label{fig: output of decision tree, depth vs tree size, guarantee}
    \includegraphics[width=0.43\linewidth]{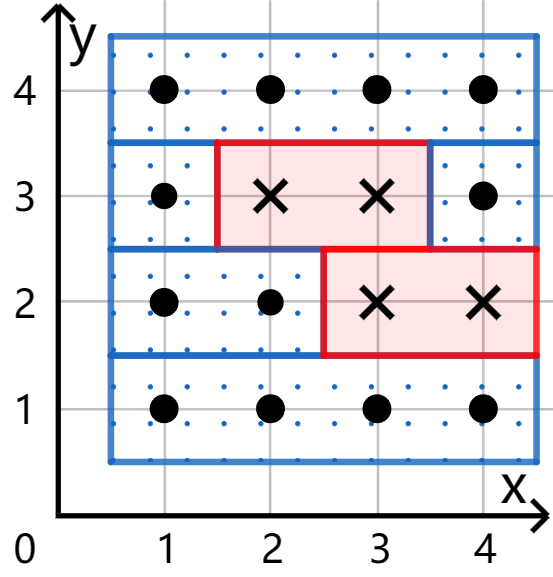}
  }
  \hspace{0.02\linewidth}
  \subfloat[The first layer of the deep tree.]{
    \label{fig: first layer of deep tree, depth vs tree size, guarantee}
    \includegraphics[width=0.43\linewidth]{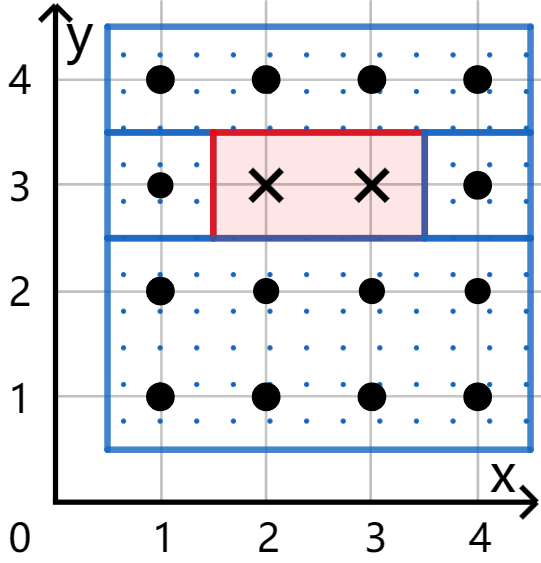}
  }
  \vspace{0.2cm}
  \subfloat[Splitting after the root node of the second layer of the deep tree.]{
    \label{fig: root of second layer of deep tree, depth vs tree size, guarantee}
    \includegraphics[width=0.43\linewidth]{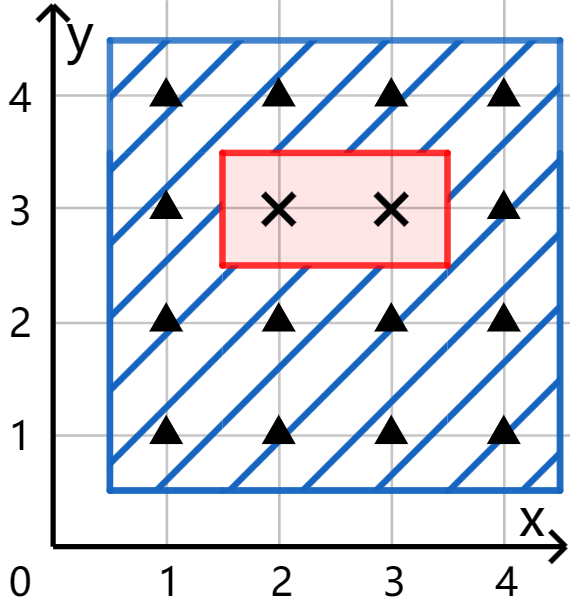}
  }
  \hspace{0.02\linewidth}
  \subfloat[The second layer of the constructed deep tree.]{
    \label{fig: rest of second layer of deep tree, depth vs tree size, guarantee}
    \includegraphics[width=0.43\linewidth]{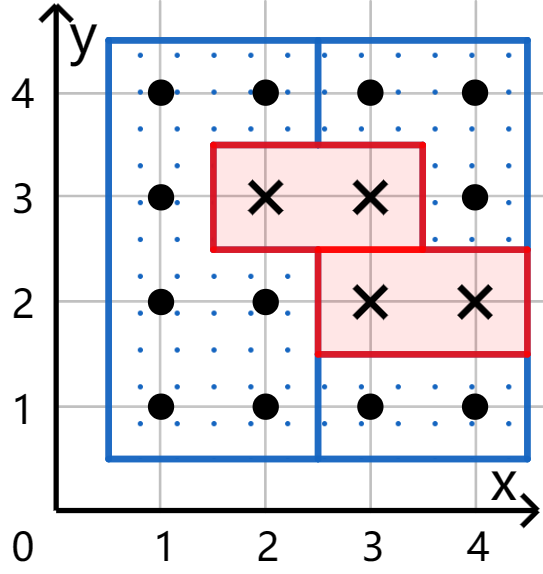}
  }
  \caption{A demonstration of expressing a decision tree using a deep tree. Circles, crosses, and triangles represent positive, negative, and unlabeled points, respectively. Rectangles indicate tree leaves.}
  \vspace{0.4cm}
\end{figure}

The main idea of the proof is that a deep tree can represent a decision tree leaf by leaf, i.e., each layer of the deep tree depicts a leaf of the decision tree and the cascade structure gathers these leaves together. Take a $2$-dimensional case as an example. Figure~\ref{fig: output of decision tree, depth vs tree size, guarantee} exhibits the output of a decision tree and we aim to find a deep tree with the same pattern as the given decision tree. It is observed that the decision tree assigns negative labels to two leaves, which motivates us to build a $2$-layer deep tree. As shown in Figure~\ref{fig: first layer of deep tree, depth vs tree size, guarantee}, the first layer of the deep tree simply divides the input space along the boundary of a positive leaf. Then this positive leaf obtains a positive pseudo label, and the other leaves receive negative ones. Figure~\ref{fig: root of second layer of deep tree, depth vs tree size, guarantee} illustrates the root node of the second layer of the deep tree, which utilizes the cascaded dimension to provide positive labels for points with positive pseudo labels and remains the rest points unlabeled. Then as pictured by Figure~\ref{fig: rest of second layer of deep tree, depth vs tree size, guarantee}, the rest nodes of the second layer directly isolate the input space along the frontier of the other positive leaf, labeling this leaf as a positive leaf and the rest ones as negative ones. One can immediately extend this construction to general input dimensions and decision trees. The number of layers of the constructed deep tree does not surpass the number of leaves of the decision tree, which possesses the same order of the approximation complexity as deep trees. Meanwhile, each layer of the deep tree just separates a hyperrectangle from the input space, which requires $O(n)$ parameters. Thus, the approximation complexity of deep trees is at most of order $n$ times the approximation complexity of decision trees.

\subsection{Depth can be more powerful than width}
\label{sec: depth vs width}

In this subsection, we compare the efficiency of depth and width.

\begin{theorem}
  \label{thm: depth vs width, separation without error}
  For any input space $\mathcal{X}$, there exist a concept $c$ and a distribution $\mathcal{D}$ over $\mathcal{X}$, such that
  \begin{equation*}
    \mathcal{C} (\mathcal{H}_\mathrm{DT} , c , \mathcal{D} , 0)
    \leqslant 10 p n 
    \quad \text{and} \quad
    \mathcal{C} (\mathcal{H}_\mathrm{F} , c , \mathcal{D} , 0)
    \geqslant p^n .
  \end{equation*}
\end{theorem}

Theorem~\ref{thm: depth vs width, separation without error} shows that there exists a classification mapping such that depth undertakes a more important role than width to efficiently express this mapping. The construction of the concept inherits the idea in Theorem~\ref{thm: depth vs tree size, separation}, i.e., the concept $c$ is the parity function. For deep trees, the proof remains the same as that of Theorem~\ref{thm: depth vs tree size, separation}. For forests, we prove the lower bound of approximation complexity by analyzing the total counts of correctly labeled points. In order to label a point properly, there should be at least one more tree assigning the correct label than the wrong label. Thus, each point contributes at least one more count to the total counts of correctly labeled points than the total counts of mistakenly labeled points. As shown in Figure~\ref{fig: leaf of tree} and its explanations, one leaf cannot cause a distinction larger than $1$ between the number of correctly labeled and mistakenly labeled points. Thus, the number of leaves is no less than the number of points in the input space, leading to $\Omega(p^n)$ approximation complexity.

\section{Experiments}\label{sec:experiments}

In this section, we verify the power of depth in experiments. Section~\ref{subsec: product distributions} introduces the construction of a product distribution, which is important to learn parity functions in simulation. Section~\ref{subsec: numerical simulation} verifies theoretical findings via simulation. Section~\ref{subsec: real world experiments} shows the advantage of depth in real-world experiments.

\subsection{Product distributions}
\label{subsec: product distributions}

As shown in Section~\ref{sec:mainresults}, generalized parity functions can be efficiently expressed by deep trees under any distribution. But from the aspect of learning, it is known that parity functions with uniform distributions are hard to learn using decision trees since there is no impurity gain at early splits~\citep{blanc21heuristics,zheng23concistency,mazumder2023sid}. This motivates us to investigate the existence of a specific distribution to demonstrate the power of depth in both approximation and learning. When considering learning tree-based models in experiments, we focus on the hypercubic input space $[p]^n$ with $p=4$ and the following probability mass function
\begin{equation*}
    p_n (\bm{x})
    = \prod_{i=1}^n f_i (x_i)
    \quad
    \forall \bm{x} \in [p]^n ,
\end{equation*}
where $f_i : [p] \rightarrow \mathbb{R}$ denotes the probability mass function of a 1-dimensional random variable defined by
\begin{equation*}
    f_i (1) = \frac{1}{b_i} , ~~
    f_i (2) = \frac{a}{b_i} , ~~
    f_i (3) = \frac{a^i}{b_i} , ~~
    f_i (4) = \frac{1}{b_i} ,
\end{equation*}
where $b_i = 2 + a + a^i$ denotes the normalization coefficient, and $a$ represents a constant. The constructed distribution is a product distribution parameterized by the constant $a$. The constant $a$ controls the extent of asymmetry and should be large enough to support efficient learning of parity functions using deep trees. In experiments, it suffices to choose $a=3$ while it fails when $a=2$.

\begin{figure}[t]
  \subfloat[The uniform distribution. The root node splits at random, and there is no impurity change.]{
    \includegraphics[width=0.43\linewidth]{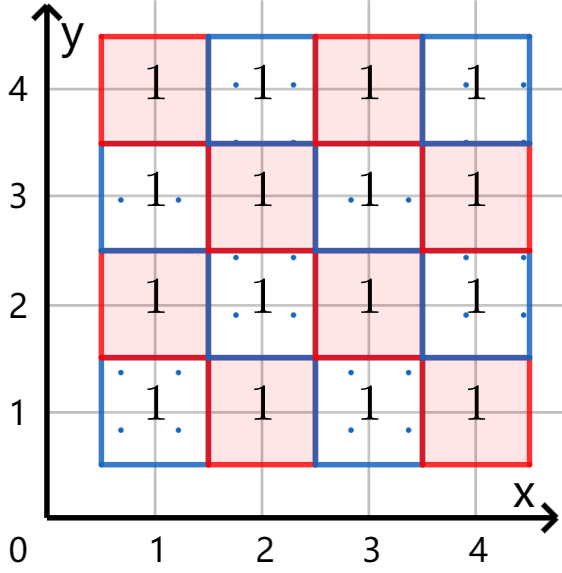}
  }
  \hspace{0.04\linewidth}
  \subfloat[The constructed product distribution. The root node splits at $x=2.5$, and impurity decreases.]{
    \includegraphics[width=0.43\linewidth]{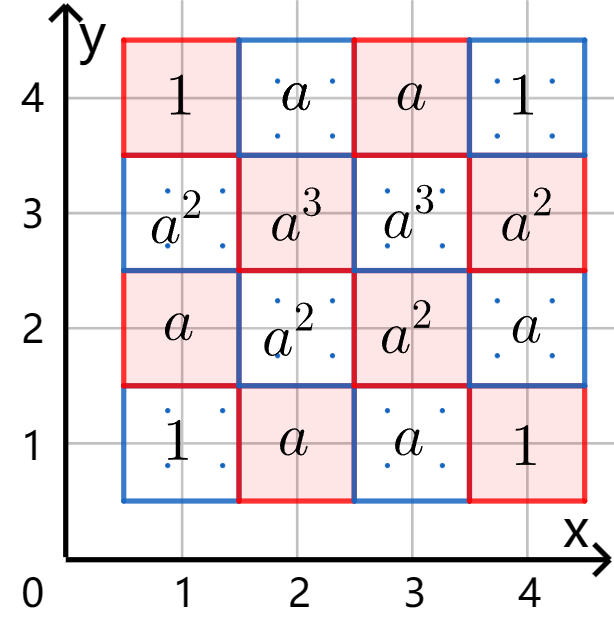}
  }
  \caption{A 2-dimensional demonstration of parity functions with the uniform distribution and the constructed product distribution. Grids with red shadow have negative labels, and grids with blue dots have positive labels. The number in each grid represents the relative magnitude of the probability mass function, and $a=3$ is a constant.}
  \label{fig: parity functions with uniform and product distributions}
  \vspace{0.4cm}
\end{figure}

\begin{figure}[ht]
    \centering
    \subfloat[The learned decision tree.]{
        \includegraphics[width=\linewidth,trim=100 130 100 130,clip]{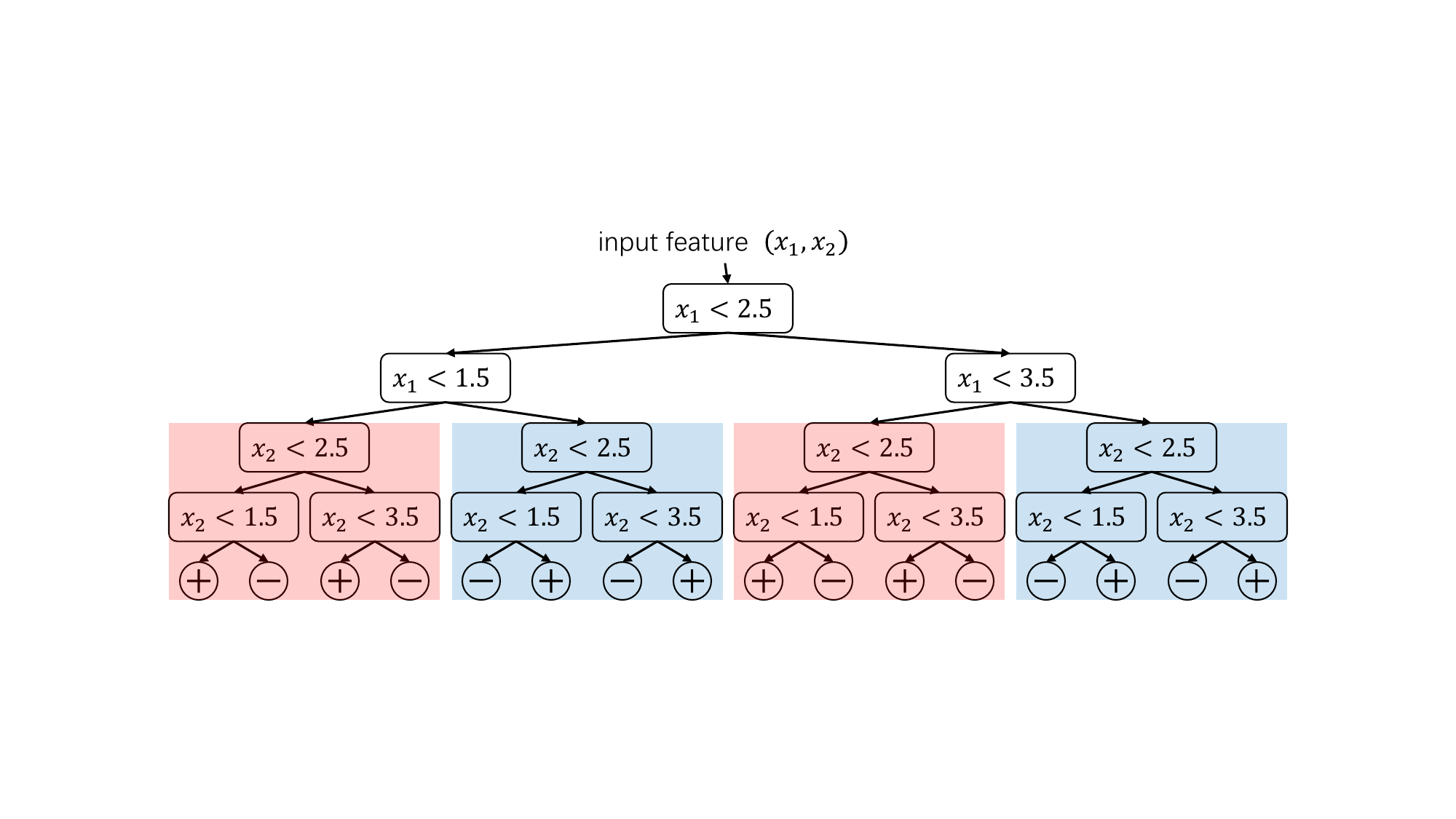}
    }
    \hspace{0.04\linewidth}
    \subfloat[The learned deep tree.]{
        \includegraphics[width=\linewidth,trim=100 150 100 150,clip]{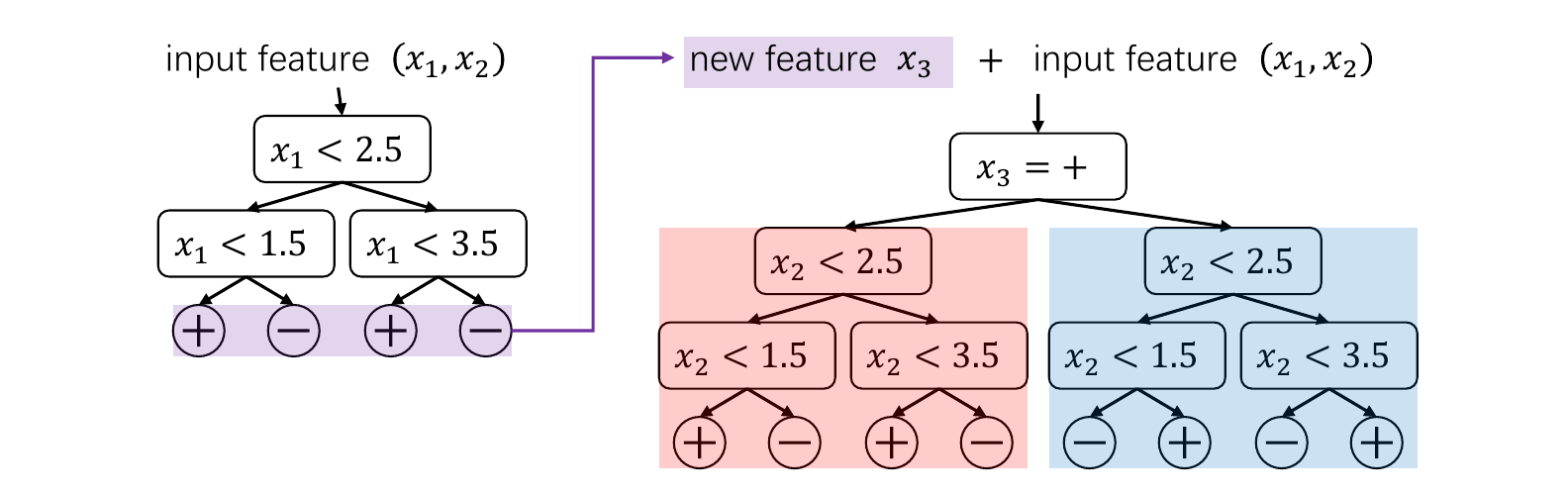}
    }
    \caption{The decision tree and deep tree learned from the 2-dimensional parity function with the constructed product distribution. Subtrees with shadows of the same color are the same. The deep tree merges identical subtrees and uses fewer leaves.}
    \label{fig: tree for parity}
    \vspace{0.4cm}
\end{figure}

\begin{figure*}[ht]
    \centering
    \includegraphics[width=0.9\textwidth]{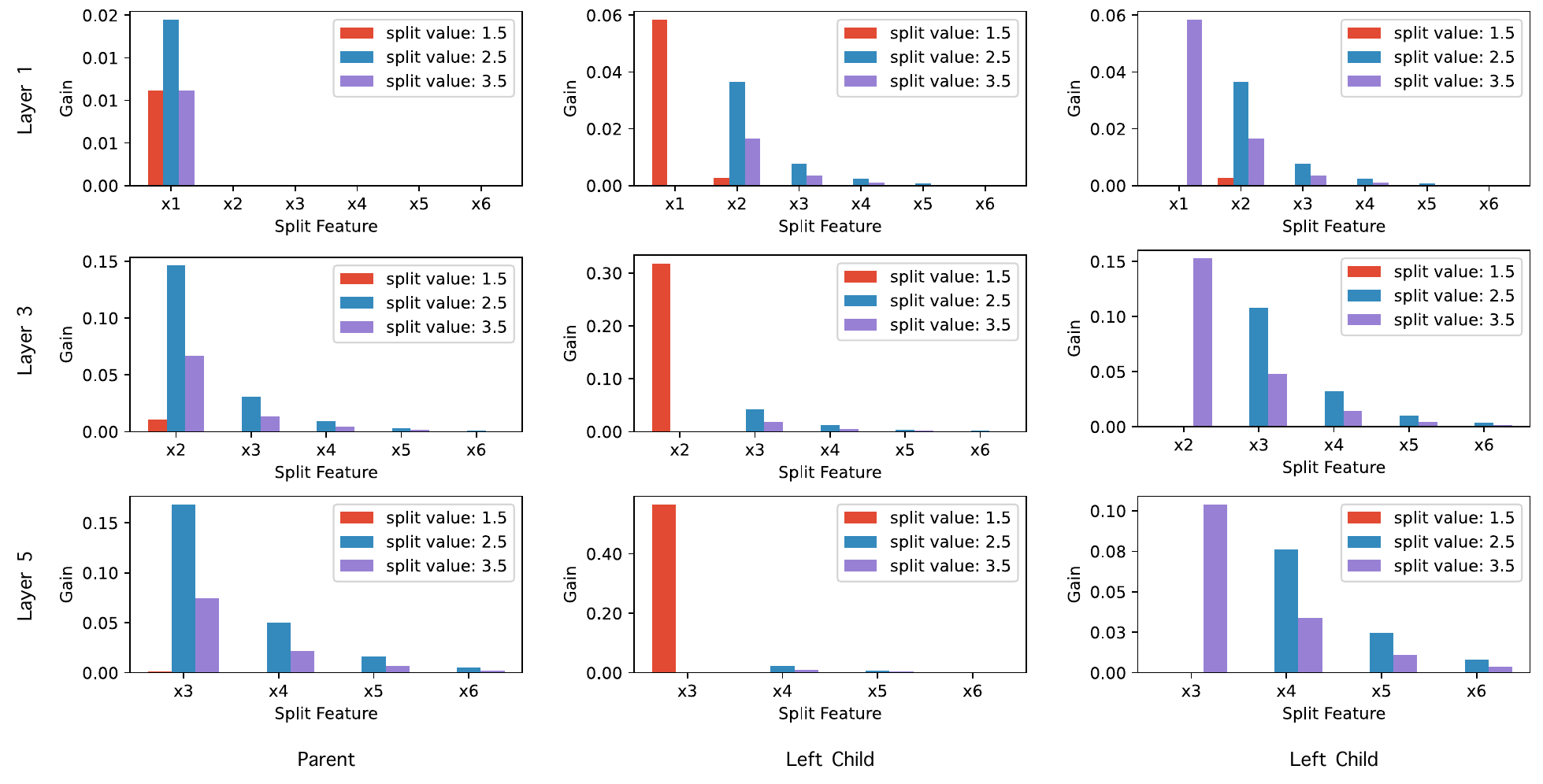}
    \caption{The purity gain in a decision tree with 6 layers when learning 6-dimensional parity functions with the constructed product distribution. The first column demonstrates gains in the first, third, and fifth layers. The second and third columns show gains in the left and right child nodes of the parent node in the first column, respectively. By the principle of maximizing purity gain, each parent node splits at the midpoint of a feature, and its child node splits at midpoints of the same feature.}
    \label{fig: purity gain of learning 6-d parity using decision tree}
    \vspace{0cm}
\end{figure*}

The intuition behind the construction is that the distribution should be highly asymmetric to meet two requirements. Firstly, the root node splits at the midpoint of a feature, and its child nodes split at the midpoint of the same feature until this feature can no longer be split. Secondly, all nodes choose the same feature as the next splitting feature and repeat the first requirement after one feature is split completely. These two requirements lead to highly symmetric decision paths and help deep trees learn parity functions efficiently.

The constructed product distribution makes learning parity functions more efficient, compared with the uniform distribution. We demonstrate 2-dimensional examples of parity functions with the uniform distribution and the constructed product distribution in Figure~\ref{fig: parity functions with uniform and product distributions}. Consider the root node of a decision tree. For the uniform distribution, the probability mass function is mirror symmetric along both $x=2.5$ and $y=2.5$. Thus, there is no impurity change to split any feature at any point, which leads to a random split at the root node. While for the constructed product distribution, the probability mass distribution is mirror symmetric only along $x=2.5$. Thus, a vertical split breaks the symmetry and brings impurity decrease. In $n$-dimensional spaces, the uniform distribution has $n$ axes of symmetry, and it takes at least $n$ layers in a decision tree to start the reduction of impurity. While the constructed product distribution only has 1 axis of symmetry since only $f_1$ is symmetric. Thus, the root node splits using $x_1$, and the impurity decreases in each layer.

Furthermore, the constructed product distribution induces symmetric decision paths in a decision tree and makes deep trees more efficient than decision trees. We demonstrate the decision tree and deep tree learned from the 2-dimensional parity function with the constructed product distribution in Figure~\ref{fig: tree for parity}. When learning using decision trees, the learned tree is symmetric in the sense that the two red subtrees are the same, and the two blue subtrees are the same. When learning using deep trees with a suitable depth, the output of the first tree, which is also the new feature in the second tree, automatically merges identical subtrees. For high-dimensional parity functions with the constructed product distribution, numerical calculation verifies that the distribution satisfies the two requirements mentioned above when the input dimension $n \leqslant 8$. Due to the limited space, we demonstrate the verification of $n=6$ in Figure~\ref{fig: purity gain of learning 6-d parity using decision tree} and provide verification of all $n \leqslant 8$ in the supplementary materials. We use Gini index as the measure of impurity and plot the purity gain in a decision tree with 6 layers when learning 6-dimensional parity functions with the constructed product distribution. As shown in the first row, the first and second layers split at midpoints of the same feature $x_1$. Then $x_1$ only takes one value in each leaf and is removed from the horizontal axis. In the third layer, the distributions in the 4 leaves are the same since all distributions are product distributions. Thus, it suffices to consider the purity gain in one leaf. We repeat this procedure and find that all nodes split at the midpoint of one feature, and nodes in even-numbered layers split at the same feature as their parent nodes. Therefore, the learned decision tree is highly symmetric, and the number of identical subtrees is exponential with respect to $n$. Then deep trees can reduce the model complexity exponentially. The numerical verification of higher dimensions is difficult since the $4^n$-dimensional probability mass matrix exceeds storage limits.

\subsection{Numerical simulation}
\label{subsec: numerical simulation}

In this subsection, we verify our theories by comparing the performance of trees, deep trees, and random forests via simulation.

\textbf{Datasets.} We randomly sample $10^6$ points from the probability mass function $p_n$. Inputs are generated by perturbing all points with random noises from the uniform distribution over $[-0.5 , 0.5]^n$. The label of an input is the same as the output of the parity function on its nearest integer lattice point., i.e.,
\begin{equation*}
    y (\bm{x})
    = c (\bm{x}_0)
    \quad \text{with} \quad
    \bm{x}_0 = \mathrm{argmin}_{\bm{x}' \in [p]^n} \Vert \bm{x}' - \bm{x} \Vert^2 ,
\end{equation*}
where $c (\bm{x}) = (-1)^{\Vert \bm{x} \Vert_1}$ is the parity function. When some coordinate of $\bm{x}$ is 0.5, which happens with 0 probability and does not impact on the performance, the minimizer is not unique, and the label is randomly chosen. These 1,000,000 pairs of inputs and labels form a dataset, of which 70\% are used as a training set, and the remaining 30\% are used as a test set.

\begin{figure*}[t]
    \centering
    \subfloat[$n=2$.]{
        \includegraphics[width=0.33\linewidth]{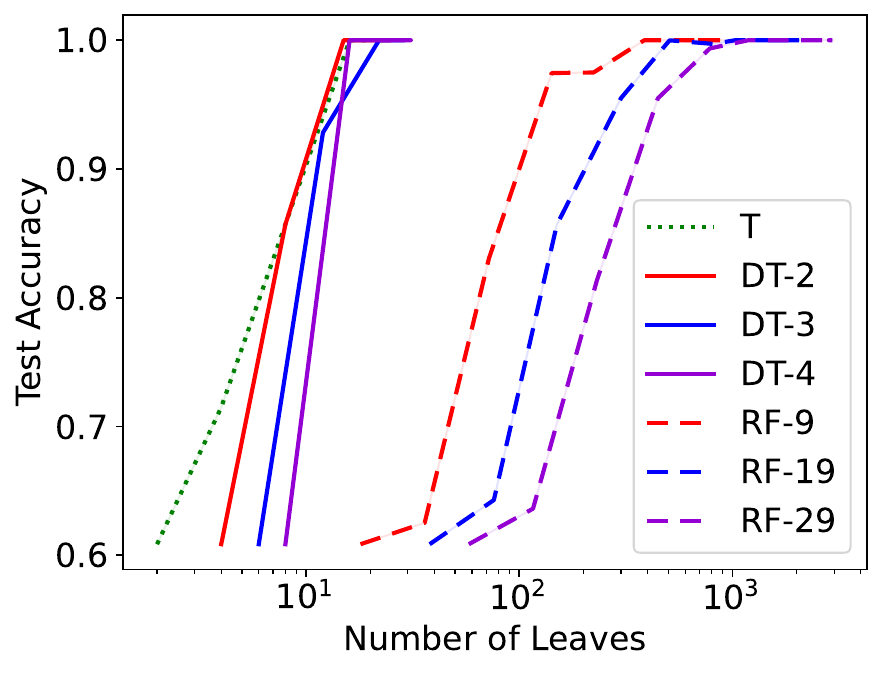}
    }
    \subfloat[$n=4$.]{
        \includegraphics[width=0.33\linewidth]{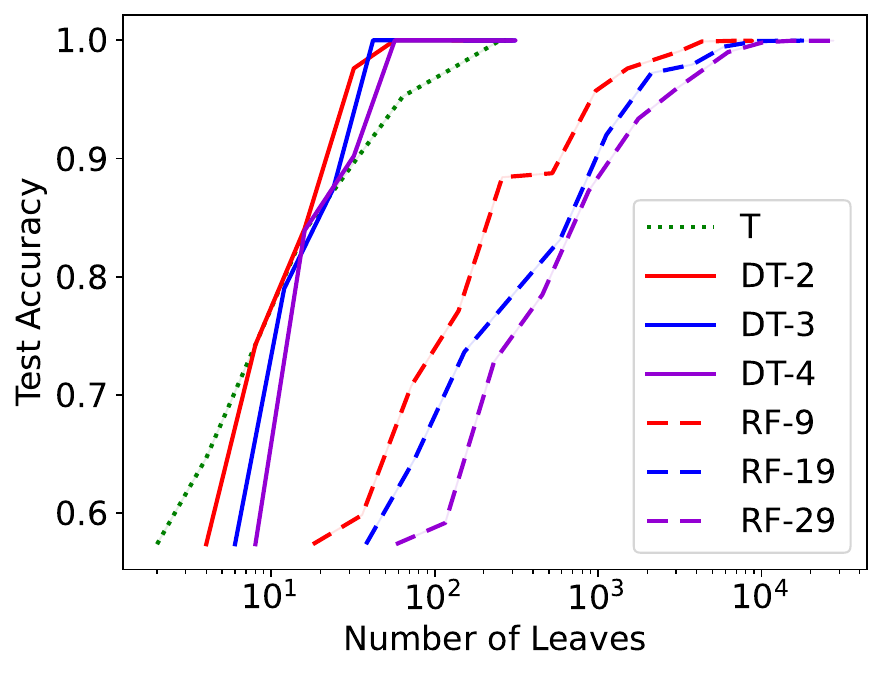}
    }
    \subfloat[$n=8$.]{
        \includegraphics[width=0.33\linewidth]{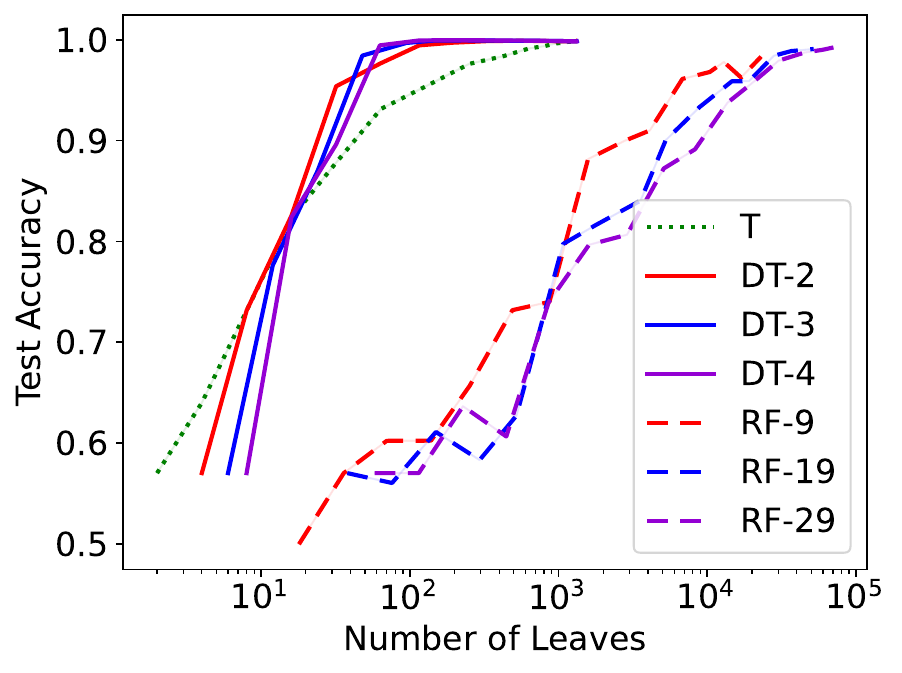}
    }
    \caption{Test accuracy of learning $n$-dimensional parity functions with the constructed product distribution using trees, deep trees of different depths, and random forests of different numbers of trees. The test accuracy is plotted as a function of total number of leaves. Deep trees can achieve the same performance using fewer leaves, which is more obvious when dealing with high-dimensional inputs.}
    \vspace{0cm}
    \label{fig: simulation experiments}
\end{figure*}

\begin{figure*}[t]
    \centering
    \subfloat[Pendigits dataset.]{
        \includegraphics[width=0.33\linewidth]{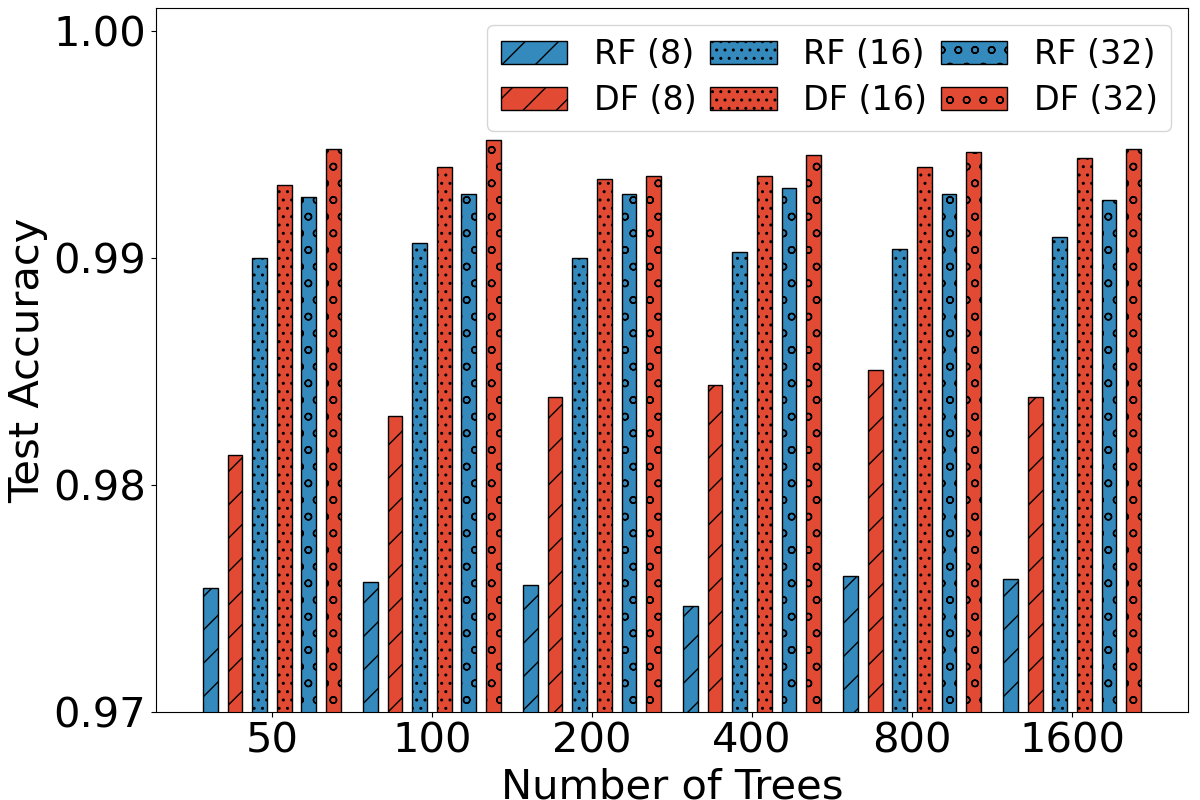}
    }
    \subfloat[Satimage dataset.]{
        \includegraphics[width=0.33\linewidth]{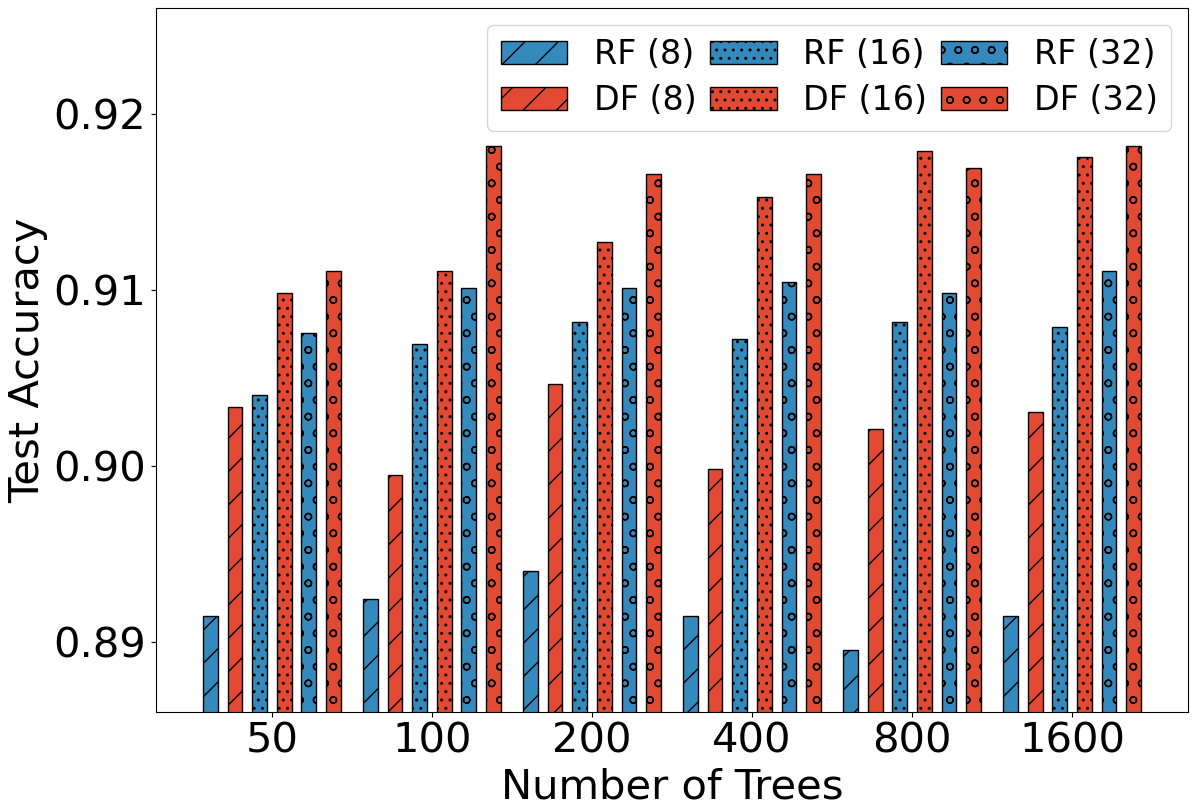}
    }
    \subfloat[Segment dataset.]{
        \includegraphics[width=0.33\linewidth]{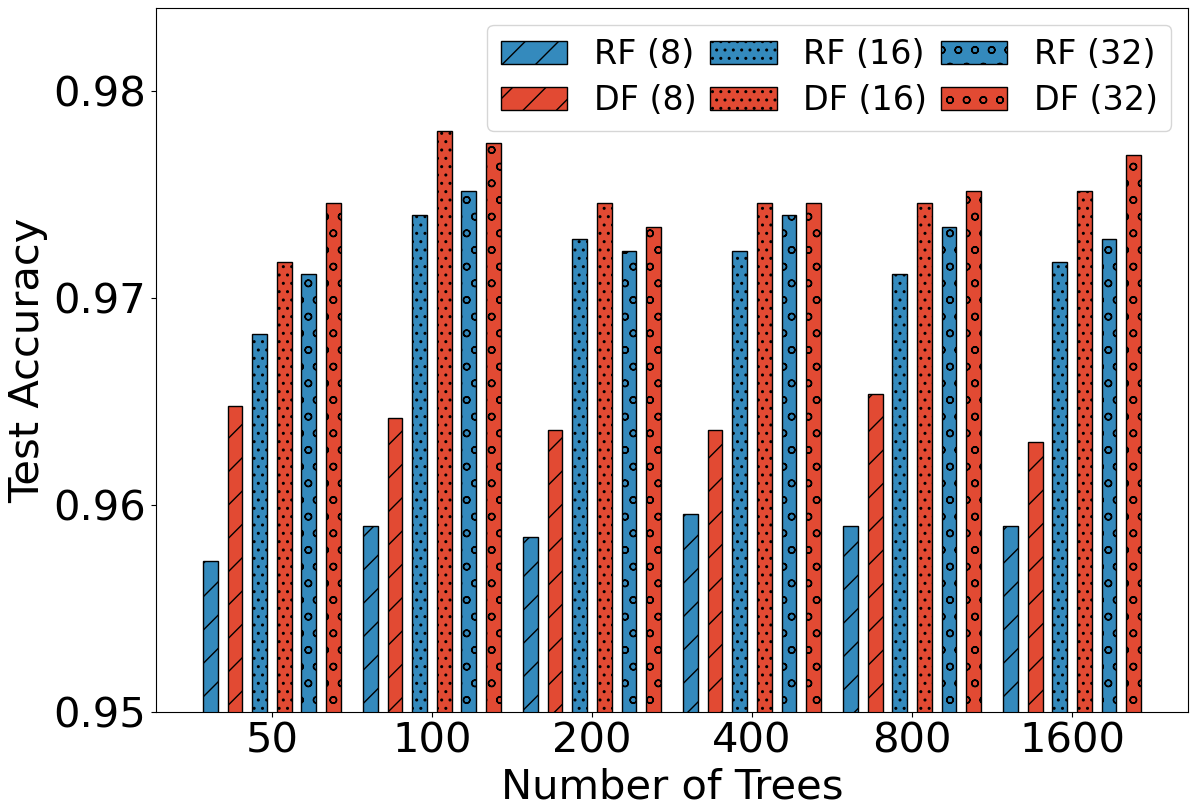}
    }
    \caption{Test accuracy of learning three benchmark datasets with random forests (RF) and deep forests (DF) of different numbers of trees and different tree sizes of base learners. `($\cdot$)' denotes the tree size of base learners. }
    \vspace{0cm}
    \label{fig: benchmark experiments}
\end{figure*}

\textbf{Training and testing.} We use the training set to train decision trees (T), deep trees with depth 2 (DT-2), 3 (DT-3), and 4 (DT-4), and random forests with 9 (RF-9), 19 (RF-19), and 29 (RF-29) trees. The maximum depth of trees in these models varies from 1 to 15. The Gini index is used as the splitting criterion. Other parameters use the default values implemented by scikit-learn~\cite{pedregosa2011scikit}. We test these models on the test set and record the test accuracy.

\textbf{Simulation results.} In Figure~\ref{fig: simulation experiments}, we plot the test accuracy with respect to the total number of leaves, which equals the summation of the number of leaves over all trees. We also provide the total number of leaves needed to achieve a test accuracy of 99\% in Table~\ref{tab: total number of leaves to learn parity}.  Deep trees achieve high accuracy with the fewest number of leaves. As the input dimension $n$ increases, deeper deep trees become more efficient, and the gap between deep trees and other models becomes larger. These results support the power of depth in theories.

It is observed that random forests require a much larger number of leaves in the simulation since there are plenty of labeled samples without label noise, which can be done with a single tree. In such a task, diversity from random forests only takes more leaves and hardly helps training. However, these results do not eliminate the importance of width and diversity in deep forests since real-world data usually has noise and a restricted number of samples. 

\begin{table}[h]
    \centering
    \vspace{0.3cm}
    \caption{The total number of leaves needed to achieve a test accuracy of 99\% using different models, where a horizontal line means failure of achieving an accuracy of 99\%, and the bold number indicates the fewest number of leaves. Deep trees use the fewest number of leaves.}
    \vspace{0.2cm}
    \begin{tabular}{cccccccc}
        \toprule
         & T & DT-2 & DT-3 & DT-4 & RF-9 & RF-19 & RF-29 \\
         \midrule
         $n=2$ & 16 & \textbf{15} & 22 & 16 & 385 & 507 & 783 \\
         $n=4$ & 251 & 57 & \textbf{42} & 57 & 3,142 & 5,918 & 6,255 \\
         $n=8$ & 633 & 116 & 95 & \textbf{63} & --- & 51,972 & 59,795 \\
         \bottomrule
    \end{tabular}
    \label{tab: total number of leaves to learn parity}
\end{table}

\subsection{Real-world experiments}
\label{subsec: real world experiments}

In this subsection, we verify our theories by comparing the performance of deep forests and random forests with different tree sizes on three real-world datasets.

\textbf{Datasets.} We select three widely used benchmark datasets of classification tasks from the UCI Machine Learning Repository \citep{uci}. The datasets vary in size: 
from 2310 to 10992 instances, from 16 to 36 features, and from 6 to 10 classes. From the literature, these datasets come pre-divided into training and testing
sets. Therefore in our experiments, we use them in their original format.

\textbf{Training and testing.} We use the training set to train random forests with width in $\{50, 100, 200, 400, 800, 1600\}$ and deep forests with depth 2 and width in $\{25, 50, 100, 200, 400, 800\}$, so that the total numbers of trees in both equal. We also set all the tree sizes in $\{8,16,32\}$ to show the influence of different sizes of trees as base learners on the ensembles. Other parameters use the default values implemented by scikit-learn~\cite{pedregosa2011scikit}. We test these models on the test set and record the test accuracy.

\textbf{Benchmark results.} We plot the test accuracy of random forests and two-layer deep forests with different tree sizes and widths in Figure~\ref{fig: benchmark experiments}. Under equivalent tree size and number of trees, deep forests consistently outperform random forests. Even when the width of random forests significantly exceeds that of deep forests, its performance struggles to match that of deep forests. These results demonstrate that depth, compared to width, provides greater efficiency.
It is observed that the tree size of base learners plays a crucial role in practice. 
While individual decision trees with larger tree sizes perform poorly, random forests and deep forests constructed from these trees outperform those built with smaller tree sizes.

\section{Conclusion}\label{sec:conclusion}

This paper provides the first comparison of tree size, width, and depth from the aspect of expressiveness. We theoretically prove that depth is more powerful than tree size and width when approximating parity functions. Experiments show that our theoretical results are valid in many objective function classes other than parity functions.
In the future, it is promising to investigate the learnability advantage~\citep{wu2023complex,xu2023over} of depth in deep forests and analyze the robustness of deep forests when dealing with noisy data.




\begin{ack}
Shen-Huan Lyu was supported by the National Natural Science Foundation of China (62306104), Jiangsu Science Foundation (BK20230949), China Postdoctoral Science Foundation (2023TQ0104), Jiangsu Excellent Postdoctoral Program (2023ZB140). Jin-Hui Wu was supported by the Program for Outstanding PhD Candidates of Nanjing University (202401A13).

\end{ack}



\bibliography{ecai}

\appendix\onecolumn

\section{Mathematical symbols}
\label{app: mathematical symbols}

In this section, we provide a table of mathematical symbols in Table~\ref{tab: mathematical symbols}.

\begin{table}[ht]
    \centering
    \caption{Mathematical symbols.}
    \label{tab: mathematical symbols}
    \vspace{0.1cm}
    \begin{tabular}{ll}
        \toprule
        Symbol & Explanation \\
        \midrule
        $c$ & target concept \\
        $\mathcal{C} (\mathcal{H} , c , \mathcal{D} , \epsilon)$ & complexity of approximating concept $c$ using hypothesis space $\mathcal{H}$ under distribution $\mathcal{D}$ with accuracy $\epsilon$ \\
        $C_{r,f} (\bm{x})$ & $(r,f)$-label-connected set of vector $\bm{x}$ \\
        $D$ & depth of deep trees \\
        $\mathcal{D}$ & distribution of input \\
        $\mathrm{dim} (\cdot)$ & dimension of a vector or a space \\
        $\mathcal{E} (h,c) , \mathcal{P} (h,c)$ & error set and proper set of hypothesis $h$ with target concept $c$ \\
        $h$ & a hypothesis \\
        $\mathcal{H}_{\mathrm{T}} , \mathcal{H}_{\mathrm{T}^\circ}$ & hypothesis space of decision trees, decision trees of restricted size \\
        $\mathcal{H}_{\mathrm{F}} , \mathcal{H}_{\mathrm{DT}}$ & hypothesis space of forests and deep trees composed of trees with restricted size \\
        $L$ & number of leaves in decision trees \\
        $M (\bm{v})$ & the mode (or majority vote) of the vector $\bm{v}$ \\
        $n$ & dimension of input \\
        $N$ & number of trees in forests \\
        $[p]$ & set $\{ 1 , 2 , \dots , p \}$ \\
        $\mathcal{P}_{\mathrm{T}}$ & parameter space of binary decision trees \\
        $\bm{v}$ & vector, whose $i$-th coordinate is $v_i$ \\
        $\mathcal{X}$ & input space $\mathcal{X} = \mathcal{X}_1 \times \mathcal{X}_2 \times \dots \times \mathcal{X}_n$ with $\mathcal{X}_i = [p]$ \\
        $\mathcal{Y}$ & output space $\mathcal{Y} = \{ -1 , +1 \}$ \\
        $\oplus$ & cascade composition $g \oplus f = g (x , f(x))$ with $f : \mathcal{X} \rightarrow \mathcal{Y}$ and $g : \mathcal{X} \times \mathcal{Y} \rightarrow \mathcal{Z}$ \\
        $\Vert \cdot \Vert_1$ & 1-norm of a vector \\
        $| \cdot |$ & absolute value of a scalar or cardinality of a set \\
        \bottomrule
    \end{tabular}
\end{table}

\section{Complete proof for Section~\ref{subsec: generalized parity functions}}
\label{app: complete proof for parity functions and product distributions}

\textbf{Proof of Lemma~\ref{lem: coincident or disjoint label connected sets}.} We prove this lemma by discussing the label connection of input vectors $\bm{x}$ and $\bm{y}$.

Case 1: vectors $\bm{x}$ and $\bm{y}$ are $(r,f)$-label-connected. Then there exist $\bm{z}_1 = \bm{x} , \bm{z}_2 , \dots , \bm{z}_n = \bm{y}$, such that $\Vert \bm{z}_{i+1} - \bm{z}_i \Vert_1 \leqslant r$ and $f(\bm{z}_{i+1}) = f(\bm{z}_i)$ holds for any $i \in [n-1]$. For any vector $\bm{w} \in C_{r,f}(\bm{y})$, there exist $\bm{z}_{n+1} = \bm{y} , \bm{z}_{n+2} , \dots , \bm{z}_{n+n'} = \bm{w}$, such that $\Vert \bm{z}_{n+i+1} - \bm{z}_{n+i} \Vert_1 \leqslant r$ and $f(\bm{z}_{n+i+1}) = f(\bm{z}_{n+i})$ holds for any $i \in [n'-1]$. Since $\bm{z}_{n+1} = \bm{z}_n = \bm{y}$, we have $\Vert \bm{z}_{n+1} - \bm{z}_n \Vert_1 \leqslant r$ and $f(\bm{z}_{n+1}) = f(\bm{z}_n)$. Thus, the vector series $\{ \bm{z}_i \}_{i=1}^{n+n'}$ satisfies $\Vert \bm{z}_{i+1} - \bm{z}_i \Vert_1 \leqslant r$ and $f(\bm{z}_{i+1}) = f(\bm{z}_i)$ for any $i \in [n+n'-1]$, i.e., vectors $\bm{z}_1 = \bm{x}$ and $\bm{z}_{n+n'} = \bm{w}$ are $(r,f)$-label-connected. Then according to Definition~\ref{def: label connected}, we have $\bm{w} \in C_{r,f}(\bm{x})$, which leads to $C_{r,f}(\bm{y}) \subset C_{r,f}(\bm{x})$. Similarly, we have $C_{r,f}(\bm{x}) \subset C_{r,f}(\bm{y})$. Therefore, we obtain $C_{r,f}(\bm{x}) = C_{r,f}(\bm{y})$.

Case 2: vectors $\bm{x}$ and $\bm{y}$ are not $(r,f)$-label-connected. Suppose that there is a vector $\bm{w} \in C_{r,f}(\bm{x}) \cap C_{r,f}(\bm{y})$. According to $\bm{w} \in C_{r,f}(\bm{x})$, there exist $\bm{z}_1 = \bm{x} , \bm{z}_2 , \dots , \bm{z}_n = \bm{w}$, such that $\Vert \bm{z}_{i+1} - \bm{z}_i \Vert_1 \leqslant r$ and $f(\bm{z}_{i+1}) = f(\bm{z}_i)$ holds for any $i \in [n-1]$. Meanwhile, $\bm{w} \in C_{r,f}(\bm{y})$ indicates that there exist $\bm{z}_{n+1} = \bm{w} , \bm{z}_{n+2} , \dots , \bm{z}_{n+n'} = \bm{y}$, such that $\Vert \bm{z}_{n+i+1} - \bm{z}_{n+i} \Vert_1 \leqslant r$ and $f(\bm{z}_{n+i+1}) = f(\bm{z}_{n+i})$ holds for any $i \in [n'-1]$. Since $\bm{z}_{n+1} = \bm{z}_n = \bm{w}$, we have $\Vert \bm{z}_{n+1} - \bm{z}_n \Vert_1 \leqslant r$ and $f(\bm{z}_{n+1}) = f(\bm{z}_n)$. Thus, the vector series $\{ \bm{z}_i \}_{i=1}^{n+n'}$ satisfies $\Vert \bm{z}_{i+1} - \bm{z}_i \Vert_1 \leqslant r$ and $f(\bm{z}_{i+1}) = f(\bm{z}_i)$ for any $i \in [n+n'-1]$, i.e., vectors $\bm{z}_1 = \bm{x}$ and $\bm{z}_{n+n'} = \bm{y}$ are $(r,f)$-label-connected, which contradicts the premise of this case. Therefore, the supposition does not hold, i.e., $C_{r,f}(\bm{x}) \cap C_{r,f}(\bm{y}) = \varnothing$.

Combining the cases above completes the proof. \hfill $\square$

~

\noindent\textbf{Proof of Lemma~\ref{lem: properties of parity functions}.} For any two inputs $\bm{x} , \bm{y} \in [p]^n$ with $\Vert \bm{x} - \bm{y} \Vert_1 = 1$, it is observed that $c(\bm{x}) \neq c(\bm{y})$ for the parity function $c(\bm{x}) = (-1)^{\Vert \bm{x} \Vert_1}$. Thus, the $(1,c)$-label-connected set of any input $\bm{x}$ satisfies $C_{1,c} (\bm{x}) = \{ \bm{x} \}$. Therefore, the cardinality of all $(1,c)$-label-connected sets satisfies
\begin{equation*}
    | \{ C_{1,c} (\bm{x}) \} |
    = | \{ \bm{x} \mid \bm{x} \in [p]^n \} |
    = p^n ,
\end{equation*}
which completes the proof of the first conclusion.

Denote by $\mathcal{R} = [p]^n$ the hypercubic input space. Define $\mathcal{R}(c,+) = \{ \bm{x} \in \mathcal{R} \mid c(\bm{x}) = +1 \}$ and $\mathcal{R}(c,-) = \{ \bm{x} \in \mathcal{R} \mid c(\bm{x}) = -1 \}$ as the positive subset and negative subset under the mapping $c$, respectively. Then according to $C_{1,c} (\bm{x}) = \{ \bm{x} \}$, we have
\begin{equation}
    \label{eq: C+ = R+ and C- = R-}
    \begin{aligned}
        | \{ C_{1,c} (\bm{x}) \mid c(\bm{x}) = 1 \} |
        = | \{ \bm{x} \mid c(\bm{x}) = 1 \} |
        = \mathcal{R} (c,+)
        \quad \text{and} \quad
        | \{ C_{1,c} (\bm{x}) \mid c(\bm{x}) = -1 \} |
        = | \{ \bm{x} \mid c(\bm{x}) = -1 \} |
        = \mathcal{R} (c,-) .
    \end{aligned}
\end{equation}
We prove the second conclusion by discussing the parity of $p$.
\begin{itemize}[itemsep=1pt,topsep=1pt]
  \item Case 1: $p$ is an even number. Define a mapping $f : \mathcal{R}(c,+) \rightarrow \mathbb{Z}^n$ as
  \begin{equation*}
      f (\bm{x}) = (x_1 , x_2 , \dots , x_{j-1} , p + 1 - x_j , x_{j+1} , \dots , x_n) ,
      \quad \forall~ \bm{x} \in \mathcal{R}(c,+) .
  \end{equation*}
  According to $x_j \in [p]$, we have $p + 1 - x_j \in [p]$, which indicates that the range of $f$ locates within the hypercube $\mathcal{R}$. Based on $\bm{x} \in \mathcal{R}(c,+)$, one knows $\Vert \bm{x} \Vert_1 \in 2 \mathbb{Z}$, which, together with $p + 1 \in 2 \mathbb{N}^+ + 1$, leads to $\Vert f(\bm{x}) \Vert_1 = \Vert \bm{x} \Vert_1 - |x_j| + |p + 1 - x_j| \in 2 \mathbb{Z} + 1$, i.e., the mapping $f$ situates its range within $\mathcal{R}(c,-)$. It is observed that the mapping $f$ is injective and surjective. Thus, the mapping $f$ is a bijection between $\mathcal{R}(c,+)$ and $\mathcal{R}(c,-)$, which leads to $| \mathcal{R}(c,+) | = | \mathcal{R}(c,-) |$.
  \item Case 2: $p$ is an odd number. If $p=1$, then the second conclusion holds obviously. For $p \geqslant 3$, define $n+2$ hypercubes as follows
  \begin{equation*}
      \begin{aligned}
          & \mathcal{R}_0 = [p-1]^n , \\
          & \mathcal{R}_j = \{ \bm{x} \in [p]^n \mid x_j = p , x_i \in [p-1] , \forall i \neq j \}
          \quad \text{for} \quad
          j \in [n] , \\
          & \text{and} \quad
          \mathcal{R}_{n+1} = \{ \bm{x} \in [p]^n \mid x_j = p , \forall j \in [n] \} .
      \end{aligned}
  \end{equation*}
  It is observed that $\mathcal{R}_0$ is an $n$-dimensional hypercube with an even side length $p-1$. Meanwhile, $\mathcal{R}_j$ is an $n-1$-dimensional hypercube with an even side length $p-1$ for any $j \in [n]$. Then one knows from the first case that $| \mathcal{R}_j (c,+) | = | \mathcal{R}_j (c,-) |$ holds for all $0 \leqslant j \leqslant n$. According to $\mathcal{R} = \mathcal{R}_0 \cup \mathcal{R}_1 \cup \dots \cup \mathcal{R}_{n+1}$, we have
  \begin{equation*}
      \begin{aligned}
          \big| | \mathcal{R}(c,+) | - | \mathcal{R}(c,-) | \big|
          &= \left| \sum_{j=0}^{n+1} | \mathcal{R}_j(c,+) | - | \mathcal{R}_j(c,-) | \right| \\
          &= \big| | \mathcal{R}_{n+1} (c,+) | - | \mathcal{R}_{n+1} (c,-) | \big| \\
          &= 1 .
      \end{aligned}
  \end{equation*}
  Then based on Eq.~\eqref{eq: C+ = R+ and C- = R-}, we obtain
  \begin{equation*}
    \big| | \{ C_{1,c} (\bm{x}) \mid c(\bm{x}) = 1 \} | - | \{ C_{1,c} (\bm{x}) \mid c(\bm{x}) = -1 \} | \big|
    = \big| | \mathcal{R}(c,+) | - | \mathcal{R}(c,-) | \big|
    = 1 .
  \end{equation*}
\end{itemize}
Combining the cases above completes the proof of the second conclusion. \hfill $\square$

~

\noindent\textbf{Proof of Lemma~\ref{lem: parity function and decision tree}.} Note that any leaf of a decision tree is a hyperrectangle. Let hyperrectangles $\mathcal{R}_1 , \mathcal{R}_2 , \dots , \mathcal{R}_L \subset \mathbb{Z}^n$ represent the leaves of the decision tree $h_\mathrm{T}$. Let $\mathcal{E}_j (h_\mathrm{T} , c) = \{ \bm{x} \in \mathcal{R}_j \mid h_\mathrm{T} (\bm{x}) \neq c (\bm{x}) \}$ indicate the error set of decision tree $h_\mathrm{T}$ within the $j$-th leaf. It is observed that a decision tree assigns the same label to all points in a leaf, i.e., either $h_\mathrm{T} |_{\mathcal{R}_j} \equiv -1$ or $h_\mathrm{T} |_{\mathcal{R}_j} \equiv +1$ holds for all $j \in [L]$. Recalling that $\mathcal{R}(c,+) = \{ \bm{x} \in \mathcal{R} \mid c(\bm{x}) = +1 \}$ and $\mathcal{R}(c,-) = \{ \bm{x} \in \mathcal{R} \mid c(\bm{x}) = -1 \}$ represent the positive subset and negative subset under the mapping $c$, respectively. Then based on $\min \{ a , b \} = (a+b)/2 - |a-b|/2$, we have
\begin{equation*}
  | \mathcal{E}_j (h_\mathrm{T} , c) |
  \geqslant \min \{ | \mathcal{R}_j (c,+) | , | \mathcal{R}_j (c,-) | \}
  = \frac{ | \mathcal{R}_j (c,+) | + | \mathcal{R}_j (c,+) | }{2} - \frac{ | | \mathcal{R}_j (c,+) | - | \mathcal{R}_j (c,+) | | }{2} .
\end{equation*}
From Lemma~\ref{lem: properties of parity functions}, $| | \mathcal{R}_j (c,+) | - | \mathcal{R}_j (c,-) | | \leqslant 1$ holds for all $j \in [L]$, which, together with $| \mathcal{R}_j (c,+) | + | \mathcal{R}_j (c,+) | = | \mathcal{R}_j |$, leads to $| \mathcal{E}_j (h_\mathrm{T} , c) | \geqslant ( | \mathcal{R}_j | - 1 ) / 2$. Summing the above inequality from $j=1$ to $L$ yields
\begin{equation*}
  | \mathcal{E} (h_\mathrm{T} , c) |
  = \sum_{j=1}^L | \mathcal{E}_j (h_\mathrm{T} , c) |
  \geqslant \sum_{j=1}^L \frac{ | \mathcal{R}_j | - 1 }{2}
  = \frac{| \mathcal{X} | - L}{2}
  = \frac{p^n - L}{2} .
\end{equation*}
According to $| \mathcal{E} (h_\mathrm{T} , c) | + | \mathcal{P} (h_\mathrm{T} , c) | = | \mathcal{X} | = p^n$, the above inequality leads to $| \mathcal{P} (h_\mathrm{T} , c) | \geqslant (p^n + L) / 2$. \hfill $\square$

\section{Omitted details for Section~\ref{sec: depth vs tree size}}

In this section, we present omitted details for Section~\ref{sec: depth vs tree size}, including proofs of Theorem~\ref{thm: depth vs tree size, separation} (in Appendix~\ref{subsec: proof of depth vs tree size, separation}) and Theorem~\ref{thm: depth vs tree size, worst case guarantee} (in Appendix~\ref{subsec: proof of depth vs tree size, worst case guarantee}).

\subsection{Proof of Theorem~\ref{thm: depth vs tree size, separation}}
\label{subsec: proof of depth vs tree size, separation}

\textbf{Proof.} Recalling the input space $\mathcal{X} = [p]^n$, define the concept $c$ as the parity function over $\mathcal{X}$, i.e.,
\begin{equation*}
  c(\bm{x}) = (-1)^{\Vert \bm{x} \Vert_1} , \quad \forall~ \bm{x} \in \mathcal{X} ,
\end{equation*}
and the distribution $\mathcal{D}$ as the uniform distribution over $\mathcal{X}$, i.e., $\Pr[ \bm{x} = \bm{x}_0 ] = 1 / p^n$ holds for any $\bm{x}_0 \in \mathcal{X}$.

Firstly, we prove the upper bound for deep trees by construction. Define the first bunch of parameters as
\begin{equation*}
    \Theta^{(1,1)} = -1
    \quad \text{and} \quad
    \Theta^{(1,q)} = \left( 1 , q-1 , n+1 , 0 , -1 , +1 , (-1)^q  \right) ,
    \quad \forall~ q \in \{ 2 , 3 , \dots , p \} .
\end{equation*}
Define corresponding deep trees as $h^{(1,q)} = h_{\Theta^{(1,q)}} \oplus \dots \oplus h_{\Theta^{(1,1)}}$. It is observed that $\dim (\Theta^{(q,1)}) \leqslant 7 = 6 \dim (\mathcal{X}) + 1$ holds for any $q \in [p]$, which indicates $h^{(1,q)} \in \mathcal{H}_\mathrm{DT}$. We claim and prove by mathematical induction that for any $q \in [p]$,
\begin{equation}
  \label{eq: the first layer deep tree, proof of depth vs tree size, separation}
  h^{(1,q)} (\bm{x})
  = \left\{
    \begin{array}{ll}
      (-1)^{x_1} , & x_1 \in [q] , \\
      (-1)^q , & \text{otherwise} ,
    \end{array}
  \right.
  \quad \text{and} \quad
  \dim (h^{(1,q)}) \leqslant  7q .
\end{equation}
\begin{itemize}
  \item Base case. The parameter $\Theta^{(1,1)} = -1$ leads to $h^{(1,1)} = -1$ and $\dim (h^{(1,1)}) = 1$, i.e., the claim holds for $q=1$.
  \item Induction. Suppose that the claim holds for $q=k$ with $k \in [p-1]$. Based on the definition of $\Theta^{(1,k+1)}$, one has
  \begin{equation*}
    h_{\Theta^{(1,k+1)}} (\bm{x} , y)
    = \left\{
      \begin{array}{ll}
        -1 , & x_1 \leqslant k \text{~and~} y \leqslant 0 , \\
        +1 , & x_1 \leqslant k \text{~and~} y > 0 , \\
        (-1)^{k+1} , & x_1 \geqslant k+1 .
      \end{array}
    \right.
  \end{equation*}
  Substitute $y$ with $h^{(1,k)} (\bm{x})$ and utilize the induction hypothesis, we have
  \begin{equation*}
    h^{(1,k+1)} (\bm{x})
    = \left\{
      \begin{array}{ll}
        -1 , & x_1 \leqslant k \text{~and~} (-1)^{x_1} \leqslant 0 , \\
        +1 , & x_1 \leqslant k \text{~and~} (-1)^{x_1} > 0 , \\
        (-1)^{k+1} , & x_1 \geqslant k+1 ,
      \end{array}
    \right.
    = \left\{
      \begin{array}{ll}
        (-1)^{x_1} , & x_1 \leqslant k , \\
        (-1)^{k+1} , & x_1 \geqslant k+1 .
      \end{array}
    \right.
  \end{equation*}
  Meanwhile, one has $\dim (h^{(1,k+1)}) = \dim (h_{\Theta^{(1,k+1)}}) + \dim (h_{1,k}) \leqslant 7 + 7k = 7(k+1)$ from the definition of $h^{(1,q)}$ and the induction hypothesis, which indicates that the claim holds for $q=k+1$.
\end{itemize}
We then define a series of bunches of parameters for $d \in \{ 2 , 3 , \dots , n \}$ and $q \in \{ 2 , 3 , \dots , p \}$ as
\begin{equation*}
  \left\{
    \begin{aligned}
      & \Theta^{(d,1)}
      = \left( n+1 , 0 , +1 , -1 \right) , \\
      & \Theta^{(d,q)}
      = \left( d , q-1 , n+1 , 0 , -1 , +1 , n+1 , 0 , +1 , -1 \right) .
    \end{aligned}
  \right.
\end{equation*}
Define corresponding deep trees as $h^{(d,q)} = h_{\Theta^{(d,q)}} \oplus \dots \oplus h_{\Theta^{(d,1)}} \oplus h^{(d-1,p)}$. It is observed that $h^{(d,q)} \in \mathcal{H}_\mathrm{DT}$ since $\dim (\Theta^{(d,q)}) \leqslant 10 \leqslant 6 \dim (\mathcal{X} \times \mathcal{Y}) + 1$ holds for all $d \in \{ 2 , 3 , \dots , n \}$ and $q \in [p]$. We claim and prove by mathematical induction that for any $d \in [n]$ and $q \in [p]$, the following holds
\begin{equation*}
  h^{(d,q)} (\bm{x})
  = \left\{
    \begin{array}{ll}
      (-1)^{x_1 + x_2 + \dots + x_d} , & x_d \in [q] , \\
      (-1)^{x_1 + x_2 + \dots + x_{d-1} + q} , & \text{otherwise} ,
    \end{array}
  \right.
  \quad \text{and} \quad
  \dim (h^{(d,q)}) \leqslant 10 p (d-1) + 10 q .
\end{equation*}
\begin{itemize}
  \item Base case. The claim holds for $d=1$ and $q \in [p]$ based on Eq.~\eqref{eq: the first layer deep tree, proof of depth vs tree size, separation}.
  \item Induction 1. Suppose that the claim holds for $d=l , q=p$ with $l \in [n-1]$. From the definition of $\Theta^{(l+1,1)}$, one has
  \begin{equation*}
    h_{\Theta^{(l+1,1)}} (\bm{x} , y)
    = \left\{
      \begin{array}{ll}
        +1 , & y \leqslant 0 , \\
        -1 , & y > 0 .
      \end{array}
    \right.
  \end{equation*}
  Substitute $y$ with $h^{(l,p)} (\bm{x})$ and utilize the induction hypothesis, we have
  \begin{equation*}
    h^{(l+1,1)} (\bm{x})
    = \left\{
      \begin{array}{ll}
        +1 , & (-1)^{x_1 + x_2 + \dots + x_l} \leqslant 0 , \\
        -1 , & (-1)^{x_1 + x_2 + \dots + x_l} > 0 ,
      \end{array}
    \right.
    = \left\{
      \begin{array}{ll}
        (-1)^{x_1 + x_2 + \dots + x_{l+1}} , & x_{l+1} \in [1] , \\
        (-1)^{x_1 + x_2 + \dots + x_l + 1} , & \text{otherwise} .
      \end{array}
    \right.
  \end{equation*}
  Meanwhile, one has $\dim (h^{(l+1,1)}) = \dim (h_{\Theta^{(l+1)}}) + \dim (h^{(l,p)}) \leqslant 4 + 10 p (l-1) + 10 p \leqslant 10 p l + 10$ from the definition of $h^{(l+1,1)}$ and the induction hypothesis, which indicates that the claim holds for $d=l+1 , q=1$.
  \item Induction 2. Suppose that the claim holds for $d=l , q=k$ with $l \in [n]$ and $k \in [p-1]$. According to the definition of $\Theta^{(l,k+1)}$, it is observed that
  \begin{equation*}
    h_{\Theta^{(l,k+1)}} (\bm{x} , y)
    = \left\{
      \begin{array}{ll}
        -1 , & x_l \leqslant k \text{~and~} y \leqslant 0 , \\
        +1 , & x_l \leqslant k \text{~and~} y > 0 , \\
        +1 , & x_l \geqslant k+1 \text{~and~} y \leqslant 0 , \\
        -1 , & x_l \geqslant k+1 \text{~and~} y > 0 .
      \end{array}
    \right.
  \end{equation*}
  Replace $y$ by $h^{(l,k)} (\bm{x})$ and apply the induction hypothesis, we have
  \begin{equation*}
    h_{\Theta^{(l,k+1)}} (\bm{x} , y)
    = \left\{
      \begin{array}{ll}
        -1 , & x_l \leqslant k \text{~and~} (-1)^{x_1 + x_2 + \dots + x_l} \leqslant 0 , \\
        +1 , & x_l \leqslant k \text{~and~} (-1)^{x_1 + x_2 + \dots + x_l} > 0 , \\
        +1 , & x_l \geqslant k+1 \text{~and~} (-1)^{x_1 + x_2 + \dots + x_{l-1} + k} \leqslant 0 , \\
        -1 , & x_l \geqslant k+1 \text{~and~} (-1)^{x_1 + x_2 + \dots + x_{l-1} + k} > 0 ,
      \end{array}
    \right.
    = \left\{
      \begin{array}{ll}
        (-1)^{x_1 + x_2 + \dots + x_l} , & x_l \leqslant k , \\
        (-1)^{x_1 + \dots + x_{l-1} + k+1} , & \text{otherwise} .
      \end{array}
    \right.
  \end{equation*}
  Furthermore, one has $\dim (h^{(l,k+1)}) = \dim (h_{\Theta^{(l,k+1)}}) + \dim (h^{(l,k)}) \leqslant 10 + 10 p (l-1) + 10 k = 10 p (l-1) + 10 (k+1)$ based on the definition of $h^{(l,k+1)}$ and the induction hypothesis. Thus, the claim holds for $d=l , q=k+1$.
\end{itemize}
By mathematical induction, the claim holds for $d=n , q=p$, i.e., $h^{(n,p)} (\bm{x}) = (-1)^{\Vert \bm{x} \Vert_1} = c(\bm{x})$ and $\dim (h^{(n,p)}) \leqslant 10 p n$. Note that $h^{(n,p)} \in \mathcal{H}_\mathrm{DT}$ is a deep tree. Therefore, we obtain
\begin{equation}
  \label{eq: approximation complexity of deep trees expressing parity function, proof of depth vs tree size, separation}
  \mathcal{C} ( \mathcal{H}_\mathrm{DT} , c , \mathcal{D} , \epsilon )
  \leqslant \mathcal{C} ( \mathcal{H}_\mathrm{DT} , c , \mathcal{D} , 0 )
  \leqslant \dim(h^{(n,p)})
  \leqslant 10 p n .
\end{equation}

Secondly, we prove the lower bound for decision trees. To achieve an error no more than $\epsilon$, it is necessary for a decision tree to compress the number of mistakenly labeled points within $\epsilon p^n$, i.e., $| \mathcal{E} (h_\mathrm{T} , c) | \leqslant \epsilon p^n$. Recalling that $L$ denotes the number of leaves of decision tree $h_\mathrm{T}$, then we have
\begin{equation*}
  \frac{p^n - L}{2}
  \leqslant | \mathcal{E} (h_\mathrm{T} , c) |
  \leqslant \epsilon p^n
  \leqslant \frac{p^n}{4} ,
\end{equation*}
where the first inequality holds from Lemma~\ref{lem: parity function and decision tree}, and the third inequality holds because of $\epsilon \leqslant 1/4$ as required in Theorem~\ref{thm: depth vs tree size, separation}. Solving the above inequality yields $L \geqslant p^n / 2$, which leads to $\mathcal{C} ( \mathcal{H}_\mathrm{T} , c , \mathcal{D} , \epsilon ) \geqslant p^n / 2$ for any $\epsilon \leqslant 1/4$.

Combining the results above completes the proof. \hfill $\square$

\subsection{Proof of Theorem~\ref{thm: depth vs tree size, worst case guarantee}}
\label{subsec: proof of depth vs tree size, worst case guarantee}

\textbf{Proof.} Let $h_\mathrm{T}$ denote the decision tree achieving the approximation complexity, i.e.,
\begin{equation}
  \label{eq: existence of tree, proof of depth vs tree size, guarantee}
  \Pr_{\bm{x} \sim \mathcal{D}} [ h_\mathrm{T}(\bm{x}) \neq c(\bm{x}) ] \leqslant \epsilon
  \quad \text{and} \quad
  \dim (h_\mathrm{T})
  = \mathcal{C} ( \mathcal{H}_\mathrm{T} , c , \mathcal{D} , \epsilon ) .
\end{equation}
It is observed that any leaf of a decision tree indicates a hyperrectangle with a coincident label, i.e., a leaf $l$ can be parameterized by $l = (a_{l,1} , b_{l,1} , a_{l,2} , b_{l,2} , \dots , a_{l,n} , b_{l,n} , y_l)$, where the first $2n$ coordinates specify the boundary of the hyperrectangle, i.e., all points in $P_l = \{ \bm{x} \in \mathcal{X} \mid \forall~ i \in [n] , a_{l,i} \leqslant x_i \leqslant b_{l,i} \}$ belong to leaf $l$, and the last one represents the label within the hyperrectangle. Let $L_+(h_\mathrm{T}) = \{ l^{(1)} , l^{(2)} , \dots , l^{(D_+)} \}$ represent the positive leaf set of the decision tree $h_\mathrm{T}$. We then construct a series of deep trees recursively as follows.
\begin{itemize}
  \item Define the first layer of the deep tree $h_{\Theta^{(1)}} = h_{\Theta_1^{(1)}}$ as
  \begin{equation}
    \label{eq: def of first layer, proof of depth vs tree size, guarantee}
    \left\{
      \begin{aligned}
        & \Theta_q^{(1)} = \left( q , a_{l^{(1)},q} - 1 , -1 , 1 , b_{l^{(1)},q} , \Theta_{q+1}^{(1)} , -1 \right) ,
        \quad \forall~ q \in [n-1] , \\
        & \Theta_n^{(1)} = \left( n , a_{l^{(1)},n} - 1 , -1 , 1 , b_{l^{(1)},n} , +1 , -1 \right) .
      \end{aligned}
    \right.
  \end{equation}
  It is observed that $\dim (h_{\Theta^{(1)}}) = \dim (\Theta^{(1)}) = 6n+1 = 6 \dim (\mathcal{X}) + 1$, which indicates that the first layer is a decision tree with restricted size, i.e., $h_{\Theta^{(1)}} \in h_\mathrm{L} (\mathcal{X} , \mathcal{Y})$. Let $h^{(1)} = h_{\Theta^{(1)}} \in \mathcal{H}_\mathrm{DT} (\mathcal{X} , \mathcal{Y})$ represent the first deep tree.
  \item For any $d \in \{ 2 , 3 , \dots , D_+ \}$, define the $d$-th layer of deep tree $h_{\Theta^{(d)}}$ as
  \begin{equation}
    \label{eq: def of dth layer, proof of depth vs tree size, guarantee}
    \left\{
      \begin{aligned}
        & \Theta^{(d)} = \left( n+1 , 0 , \Theta_1^{(d)} , +1 \right) , \\
        & \Theta_q^{(d)} = \left( q , a_{l^{(d)},q} - 1 , -1 , 1 , b_{l^{(d)},q} , \Theta_{q+1}^{(d)} , -1 \right) ,
        \quad \forall~ q \in [n-1] , \\
        & \Theta_n^{(d)} = \left( n , a_{l^{(d)},n} - 1 , -1 , 1 , b_{l^{(d)},n} , +1 , -1 \right) .
      \end{aligned}
    \right.
  \end{equation}
  It is observed that $\dim (h_{\Theta^{(d)}}) = \dim (\Theta^{(d)}) = 6n+4 \leqslant 6 \dim (\mathcal{X} \times \mathcal{Y}) + 1$, which indicates that the $d$-th layer is a decision tree with restricted size, i.e., $h_{\Theta^{(d)}} \in h_\mathrm{L} (\mathcal{X} \times \mathcal{Y} , \mathcal{Y})$. Define $h^{(d)} = h_{\Theta^{(d)}} \oplus h^{(d-1)}$ represent the $d$-th deep tree. Using mathematical induction, one can prove that $h^{(d)} \in \mathcal{H}_\mathrm{DT} (\mathcal{X} , \mathcal{Y})$ is a restricted-tree-size deep tree.
\end{itemize}
We claim and prove by mathematical induction that for any $q \in [n]$, the following holds
\begin{equation}
  \label{eq: output of the first layer, proof of depth vs tree size, guarantee}
  h_{\Theta_q^{(1)}} (\bm{x}) = \left\{
    \begin{array}{ll}
      +1 , & a_{l^{(1)},r} \leqslant x_r \leqslant b_{l^{(1)},r} \text{~for all~} r \in \{ q , q+1 , \dots , n \} , \\
      -1 , & \text{otherwise} .
    \end{array}
  \right.
\end{equation}
\begin{itemize}
  \item Base case. For $q=n$, the claim holds directly from the definition of $\Theta_n^{(1)}$ in Eq.~\eqref{eq: def of first layer, proof of depth vs tree size, guarantee}.
  \item Induction. Suppose that the claim holds for $q=k$ with $k \in \{ n , n-1 , \dots , 2 \}$. Then we have
  \begin{equation*}
    \begin{aligned}
      h_{\Theta_{k-1}^{(1)}} (\bm{x})
      & = \left\{
        \begin{array}{ll}
          h_{\Theta_k^{(1)}} (\bm{x}) , & a_{l^{(1)},k-1} \leqslant x_{k-1} \leqslant b_{l^{(1)},k-1} , \\
          -1 , & \text{otherwise} ,
        \end{array}
      \right. \\
      & = \left\{
        \begin{array}{ll}
          +1 , & a_{l^{(1)},r} \leqslant x_r \leqslant b_{l^{(1)},r} \text{~for all~} r \in \{ k-1 , k , \dots , n \}, \\
          -1 , & \text{otherwise} ,
        \end{array}
      \right.
    \end{aligned}
  \end{equation*}
  where the first equality holds based on the definition of $\Theta_{k-1}^{(1)}$ in Eq.~\eqref{eq: def of first layer, proof of depth vs tree size, guarantee}, and the second one holds according to the induction hypothesis. Thus, the claim holds for $q=k-1$.
\end{itemize}
It is observed that the definition of $\Theta_1^{(d)}$ is the same as that of $\Theta_1^{(1)}$ except all superscripts. Thus, similar conclusion holds for general layers as Eq.~\eqref{eq: output of the first layer, proof of depth vs tree size, guarantee}. Recalling that $P_l$ denotes the set of all points belonging to the leaf $l$, setting $q=1$ leads to
\begin{equation}
  \label{eq: output of auxiliary deep tree, proof of depth vs tree size, guarantee}
  h_{\Theta_1^{(d)}} (\bm{x}) = \left\{
    \begin{array}{ll}
      +1 , & \bm{x} \in P_{l^{(d)}} , \\
      -1 , & \text{otherwise} .
    \end{array}
  \right.
\end{equation}
Define $P_{\leqslant d} = P_{l^{(1)}} \cup P_{l^{(2)}} \cup \dots \cup P_{l^{(d)}}$ as the union of all points in the first $d$ leaves. We claim and prove by mathematical induction that for any $d \in [D_+]$, the following holds
\begin{equation}
  \label{eq: output of deep tree, proof of depth vs tree size, guarantee}
  h^{(d)} (\bm{x}) = \left\{
    \begin{array}{ll}
      +1 , & \bm{x} \in P_{\leqslant d} , \\
      -1 , & \bm{x} \notin P_{\leqslant d} .
    \end{array}
  \right.
  \quad \text{and} \quad
  \dim (h^{(d)}) = (6n+4)d - 3 .
\end{equation}
\begin{itemize}
  \item Base case. For $d=1$, the claim holds directly from Eq.~\eqref{eq: output of auxiliary deep tree, proof of depth vs tree size, guarantee} and $\dim (\Theta^{(1)}) = 6n+1$.
  \item Induction. Suppose that the claim holds for $d=k$ with $d \in [n-1]$. Then we have
  \begin{equation*}
    h^{(k+1)} (\bm{x})
    = h_{\Theta^{(k+1)}} ( \bm{x} , h^{(k)} (\bm{x}) )
    = \left\{
      \begin{array}{ll}
        h_{\Theta^{(k+1)}} ( \bm{x} , +1 ) , & \bm{x} \in P_{\leqslant k} , \\
        h_{\Theta^{(k+1)}} ( \bm{x} , -1 ) , & \bm{x} \notin P_{\leqslant k} ,
      \end{array}
    \right.
    = \left\{
      \begin{array}{ll}
        +1 , & \bm{x} \in P_{\leqslant k} , \\
        h_{\Theta_1^{(k+1)}} ( \bm{x} ) , & \bm{x} \notin P_{\leqslant k} ,
      \end{array}
    \right.
  \end{equation*}
  where the second equality holds from the induction hypothesis, the third equality holds based on the definition of $\Theta^{(k+1)}$ in Eq.~\eqref{eq: def of dth layer, proof of depth vs tree size, guarantee}, and the input of $h_{\Theta_1^{(k+1)}}$ omits the unused $(n+1)$-th dimension in the right side of the last equality. Then based on Eq.~\eqref{eq: output of auxiliary deep tree, proof of depth vs tree size, guarantee} and the definition of $P_{\leqslant k}$, one has
  \begin{equation*}
    h^{(k+1)} (\bm{x})
    = \left\{
      \begin{array}{ll}
        +1 , & \bm{x} \in P_{\leqslant k} , \\
        +1 , & \bm{x} \notin P_{\leqslant k} \text{~and~} \bm{x} \in P_{l^{(k+1)}} , \\
        -1 , & \bm{x} \notin P_{\leqslant k} \text{~and~} \bm{x} \notin P_{l^{(k+1)}} ,
      \end{array}
    \right.
    = \left\{
      \begin{array}{ll}
        +1 , & \bm{x} \in P_{\leqslant k+1} , \\
        -1 , & \bm{x} \notin P_{\leqslant k+1} .
      \end{array}
    \right.
  \end{equation*}
  Meanwhile, we have $\dim (h^{(k+1)}) = \dim (\Theta^{(k+1)}) + \dim (h^{(k)}) = 6n+4 + (6n+4)k - 3 = (6n+4)(k+1) - 3$, which indicates that the claim holds for $d=k+1$.
\end{itemize}
Set $d = D_+$ in Eq.~\eqref{eq: output of deep tree, proof of depth vs tree size, guarantee}, then one has
\begin{equation}
  \label{eq: construct of positive deep tree, proof of depth vs tree size, guarantee}
  h^{(D_+)} (\bm{x})
  = \left\{
    \begin{array}{ll}
      +1 , & \bm{x} \in P_{\leqslant D_+} , \\
      -1 , & \text{otherwise} ,
    \end{array}
  \right.
  = h_\mathrm{T} (\bm{x})
  \quad \text{and} \quad
  \dim (h^{(D_+)}) = (6n+4)D_+ - 3 .
\end{equation}
Similarly, we can construct a series of deep trees using the negative leaf set of the decision tree $h_\mathrm{T}$ and prove that
\begin{equation}
  \label{eq: construct of negative deep tree, proof of depth vs tree size, guarantee}
  h^{(D_-)} (\bm{x})
  = h_\mathrm{T} (\bm{x})
  \quad \text{and} \quad
  \dim (h^{(D_-)}) = (6n+4)D_- - 3 ,
\end{equation}
where $D_-$ represents the size of the negative leaf set. Note that the decision tree $h_\mathrm{T}$ owns $D_+ + D_-$ leaves, which indicates $\dim (h_\mathrm{T}) = 3 (D_+ + D_-) - 2$. From Eqs.~\eqref{eq: construct of positive deep tree, proof of depth vs tree size, guarantee} and~\eqref{eq: construct of negative deep tree, proof of depth vs tree size, guarantee}, there exists a deep tree $h_\mathrm{DT} \in \mathcal{H}_\mathrm{DT} (\mathcal{X} , \mathcal{Y})$, s.t.
\begin{equation}
  \label{eq: relation between deep tree and tree, proof of depth vs tree size, guarantee}
  h_\mathrm{DT} = h_\mathrm{T}
  \quad \text{and} \quad
  \dim (h_\mathrm{DT})
  = (6n+4) \min \{ D_+ , D_- \} - 3
  \leqslant (3n+2) ( D_+ + D_- ) - 3
  \leqslant (4n+1) \dim (h_\mathrm{T}) .
\end{equation}
Combining Eqs.~\eqref{eq: existence of tree, proof of depth vs tree size, guarantee} and~\eqref{eq: relation between deep tree and tree, proof of depth vs tree size, guarantee}, we conclude that
\begin{equation*}
  \Pr_{\bm{x} \sim \mathcal{D}} [ h_\mathrm{DT}(\bm{x}) \neq c(\bm{x}) ] \leqslant \epsilon
  \quad \text{and} \quad
  \mathcal{C} ( \mathcal{H}_\mathrm{DT} , c , \mathcal{D} , \epsilon )
  \leqslant \dim (h_{\mathrm{DT}})
  \leqslant (4n+1) \dim (h_{\mathrm{T}})
  = (4n+1) \mathcal{C} ( \mathcal{H}_\mathrm{T} , c , \mathcal{D} , \epsilon ) ,
\end{equation*}
which completes the proof. \hfill $\square$

\section{Omitted details for Section~\ref{sec: depth vs width}}

In this section, we present omitted details for Section~\ref{sec: depth vs width}, including the proof of Theorem~\ref{thm: depth vs width, separation without error} (in Appendix~\ref{subsec: proof of depth vs width, separation}).

\subsection{Proof of Theorem~\ref{thm: depth vs width, separation without error}}
\label{subsec: proof of depth vs width, separation}

\textbf{Proof.} Define the concept $c$ as the parity function over $\mathcal{X}$, i.e.,
\begin{equation*}
  c(\bm{x}) = (-1)^{\Vert \bm{x} \Vert_1} , \quad \forall~ \bm{x} \in \mathcal{X} ,
\end{equation*}
and the distribution $\mathcal{D}$ as the uniform distribution over $\mathcal{X}$, i.e., $\Pr[ \bm{x} = \bm{x}_0 ] = 1 / p^n$ holds for any $\bm{x}_0 \in \mathcal{X}$. Note that the concept $c$ and the distribution $\mathcal{D}$ remain the same as those in the proof of Theorem~\ref{thm: depth vs tree size, separation}. Thus, the conclusion and the proof about deep trees keep the same as that in Eq.~\eqref{eq: approximation complexity of deep trees expressing parity function, proof of depth vs tree size, separation}. The rest contents are devoted to the study of forests.

Let $h_\mathrm{F} \in \mathcal{H}_\mathrm{F}$ represent a forest expressing the parity function without error, i.e., $h_\mathrm{F} (\bm{x}) = c (\bm{x})$ holds for all $\bm{x} \in \mathcal{X}$. Suppose that $h_\mathrm{F}$ integrates $N$ restricted-size decision trees, namely $h_\mathrm{F} (\bm{x}) = M ( h_{\Theta_1}(\bm{x}) , h_{\Theta_2}(\bm{x}) , \dots , h_{\Theta_N}(\bm{x})) $ with $h_{\Theta_1} , h_{\Theta_2} , \dots , h_{\Theta_N} \in \mathcal{H}_\mathrm{T}$. Denote by $l_1 , l_2 , \dots , l_N$ the number of leaves of decision trees $h_{\Theta_1} , h_{\Theta_2} , \dots , h_{\Theta_N}$, respectively. Recalling that the proper set $\mathcal{P} (h , c) = \{ \bm{x} \in \mathcal{X} \mid h(\bm{x}) = c(\bm{x}) \}$ depicts the set of correctly labeled points, then Lemma~\ref{lem: parity function and decision tree} indicates $| \mathcal{P} (h_j,c) | \leqslant (p^n + l_j) / 2$ for all $j \in [N]$. Then the total count of correctly labeled points equals
\begin{equation}
  \label{eq: upper bound of P, proof of depth vs width, separation without error}
  P
  = \sum_{j=1}^N | \mathcal{P} (h_j,c) |
  \leqslant \sum_{j=1}^N \frac{p^n + l_j}{2}
  = \frac{N p^n}{2} + \frac{l_1 + l_2 + \dots + l_N}{2} .
\end{equation}
Note that the forest $h_\mathrm{F}$ assigns the correct label to a point $\bm{x}$ with probability $1$ if and only if at least $(N+1) / 2$ decision trees among $h_{\Theta_1} , h_{\Theta_2} , \dots , h_{\Theta_N}$ labels the point $\bm{x}$ correctly. Thus, the total count of correctly labeled points $P$ satisfies $P \geqslant (N+1) p^n / 2$, which, together with Eq.~\eqref{eq: upper bound of P, proof of depth vs width, separation without error}, indicates $l_1 + l_2 + \dots + l_N \geqslant p^n$. Therefore, we obtain
\begin{equation*}
  \mathcal{C} (\mathcal{H}_\mathrm{F} , c , \mathcal{D} , 0)
  \geqslant l_1 + l_2 + \dots + l_N
  \geqslant p^n ,
\end{equation*}
which completes the proof. \hfill $\square$


\end{document}